\documentclass{article} % For LaTeX2e
\usepackage{colm2024_conference}

\usepackage{tabularx}
\usepackage{booktabs}
\usepackage{listings}
\usepackage{microtype}
\usepackage{hyperref}
\usepackage{url}
\usepackage{booktabs}
\usepackage{float}
\usepackage{xspace}
\usepackage{graphicx} % Import the graphicx package
\usepackage[normalem]{ulem}
\useunder{\uline}{\ul}{}
\usepackage{amsmath}
\usepackage{multirow}
\usepackage{url}
\usepackage{caption}
\usepackage{longtable}
\usepackage{inconsolata}

\definecolor{aqua}{cmyk}{0.91, 0, 0.09, 0.36}

\usepackage{xspace}
\NewDocumentCommand\globetitle{}{
    \includegraphics[scale=0.005]{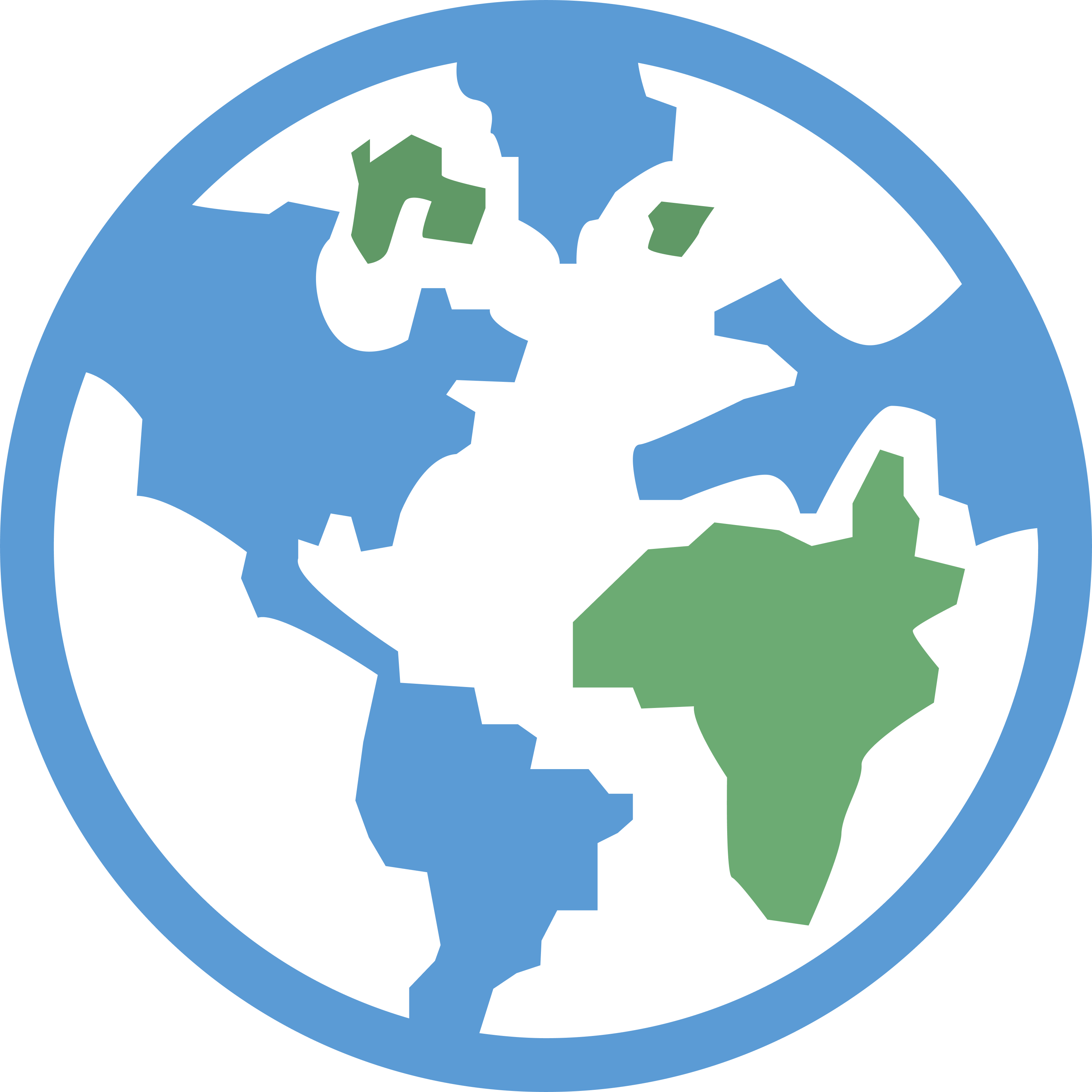}
}
\title{\globetitle\textbf{CultureBank\xspace}: An Online Community-Driven Knowledge Base Towards Culturally Aware Language Technologies}

\author{Weiyan Shi\textsuperscript{1}, Ryan Li\textsuperscript{1}, Yutong Zhang\textsuperscript{1}, Caleb Ziems\textsuperscript{1}, Chunhua Yu\textsuperscript{1}, \\ 
\textbf{Raya Horesh\textsuperscript{2}, Rogério Abreu de Paula\textsuperscript{2}, Diyi Yang\textsuperscript{1}} \\
Stanford University\textsuperscript{1}, IBM Research \textsuperscript{2}\\
\texttt{\{weiyans,lansong,yutongz7,cziems,syu03\}@stanford.edu} \\
\texttt{rhoresh@us.ibm.com, ropaula@br.ibm.com, diyiy@stanford.edu}
}

\newcommand{\dataname}{\textit{CultureBank}\xspace}

\colmfinalcopy % Uncomment for camera-ready version, but NOT for submission.
\begin{document}

\maketitle
\vspace{-1em}
\begin{abstract}
% As LLMs have more interaction with users from all over the world, it's increasingly important that they become culturally-aware. 
To enhance language models' cultural awareness, we design a generalizable pipeline to construct cultural knowledge bases from different online communities on a massive scale. With the pipeline,  we construct \dataname, a knowledge base built upon users' self-narratives with 12K  cultural descriptors sourced from TikTok and 11K from Reddit. Unlike previous cultural knowledge resources, \dataname contains diverse views on cultural descriptors to allow flexible interpretation of cultural knowledge, and contextualized cultural scenarios to help grounded evaluation. With \dataname, we evaluate different LLMs' cultural awareness, and identify areas for improvement. We also fine-tune a language model on \dataname: experiments show that it achieves better performances on two downstream cultural tasks in a zero-shot setting. Finally, we offer recommendations based on our findings for future culturally aware language technologies.\footnote{We release the \href{https://huggingface.co/datasets/SALT-NLP/CultureBank}{\color{blue}\dataname dataset}, code, and models at  \href{https://github.com/SALT-NLP/CultureBank}{\color{blue}\texttt{github.com/SALT-NLP/CultureBank}}.
} \footnote{Our project page is at \href{https://culturebank.github.io/}{\color{blue}\texttt{culturebank.github.io}}.}
\end{abstract}
\vspace{-1em}
\section{Introduction}
\vspace{-0.5em}
% motivation on why we need such a knowledge base

\begin{figure}[ht]
\centering
\includegraphics[width=\textwidth]{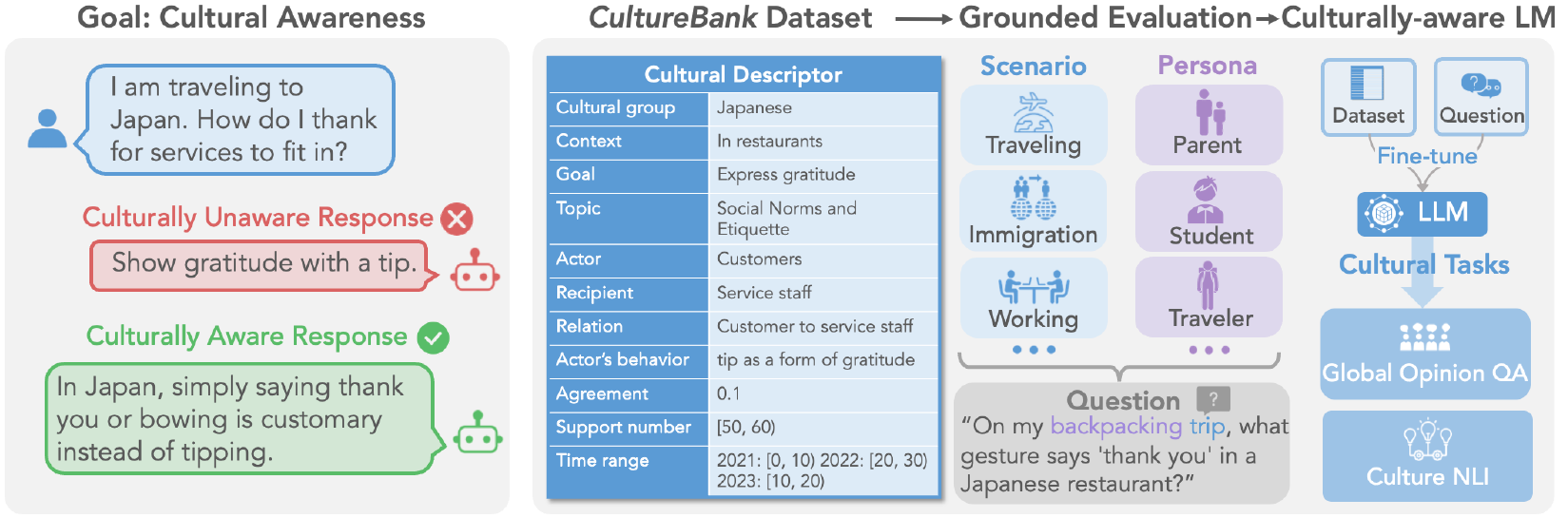}
\caption{Overview. Our goal is culturally-aware language technologies. To do so, we develop a pipeline and construct \dataname with structured cultural descriptors. Each descriptor comes with a grounded scenario, persona, and question to help evaluate LLMs. We fine-tune a model on \dataname and improve its performance on two cultural tasks.}
\label{fig:fig0}
\end{figure}
\vspace{-1em}
\begin{quote}
    {
    ``\textit{Globally, people express pride, celebrate, and respect cultural diversity, while acknowledging and working towards reducing cultural bias}'' \\\phantom{abc}--- \textbf{\dataname} 
    %\\\phantom{abcdef}inventor of the World Wide Web
    }
\end{quote}

Large Language Models (LLMs) have become instrumental in various applications  to interact with diverse user populations, such as in recommender systems \citep{li2023prompt, fan2023recommender} and customer service \citep{pandya2023automating}. %However, as these models are predominantly trained on Western-centric data that reflects these values and behaviors, they often mirror Western-centric perspectives \citep{santurkar2023whose, durmus2023towards}. Such
However, these models often mirror Western-centric perspectives \citep{santurkar2023whose, durmus2023towards}, as they are predominantly trained on data that reflect these values and behaviors. Such cultural bias can lead to unintended consequences \citep{ryan2024unintended}, e.g., reinforcing stereotypes, alienating non-Western users, hindering global deployment and so on. Therefore, it becomes increasingly important to develop language technologies that are aware of diverse cultures. 
% culturally aware language technologies. % (as shown in the dialogue in Figure~\ref{fig:fig0}). %, to present more culturally informed responses like in Figure~\ref{fig:fig0}, %, that can present more culturally informed responses like in Figure~\ref{fig:fig0}. 
% Towards more inclusive, we need to  ranging from the reinforcement of stereotypes, barrier to global deployment to the alienation of non-Western users,  when these technologies interact with diverse populations, underscoring the importance of developing language technologies that are culturally aware.

To enhance LLMs' culture\footnote{We acknowledge that culture is a broad concept, and prior work has attempted to operationalize culture via different proxies. We use \emph{culture} to refer to the knowledge shared by a relatively large group of people from different backgrounds about their shared beliefs, practices and behaviors.} awareness,  existing studies have developed cultural knowledge databases to represent culture-related knowledge and norms, but they have several limitations. (1) They often rely on formal knowledge sources like Wikipedia and online articles \citep{candle2023,fung2024massively}, which miss the rich, evolving and long-tailed cultural nuances experienced by local communities. (2) Secondly, these methods tend to present cultural knowledge in an assertive manner \citep{candle2023,fung2024massively, yin2022geomlama}, failing to capture the fact that cultural practices and values %are highly contextualized and 
can vary among individuals within the same cultural group. 
% \caleb{This is true, but ours does the same thing, right? Maybe instead, point out that prior work is top-down instead of bottom-up here. Or maybe make that point more explicit in point \#1 and then highlight something else like how CultureBank aggregates more voices than Wikipedia as used in CultureAtlas, or that it covers more countries than NormSage, or that it isn't strictly limited to geo-defined culture groups like others, or that it expands beyond the CANDLE categories of clothes,drink,food,tradition,rituals.} \wyshi{we have agreement levels for avoid being assertive}
(3) Besides, their evaluation methods often rely on classification tasks and question answering \citep{naous2023having,afina2024can, shafayat2024multi}, which is very different from how LLMs are deployed in the real world and hence cannot reflect their cultural awareness in practice.  %reflect the complex ways in which LLMs are utilized in real-world settings.%, and without incorporating the dynamic, evolving nature of culture, these databases often lack the ability to capture long-tailed, less common cultural phenomena.

To tackle these challenges, we utilize online communities where  people share their cultural experiences, and develop a bottom-up approach to process noisy self-narratives on a massive scale. Using this pipeline, we develop \dataname, a cultural knowledge base with 12K \emph{cultural descriptors} sourced from TikTok (Figure~\ref{fig:fig0} shows one example). %Besides, to address the limitation on assertiveness, we cluster comments on similar cultural practices, to calculate an agreement level for every cultural descriptor for inclusive cultural understanding.
Besides, to address the limitation on assertiveness, we gather diverse views on similar cultural practices, and 
calculate an agreement level to enable inclusive cultural understanding.
% towards certain cultural practice, to address the limitation on assertiveness.   Each cultural descriptor has an agreement level to enable inclusive cultural understanding, and time ranges to allow potential temporal cultural analysis. 
Moreover, to facilitate contextualized evaluation on LLMs' cultural awareness, we provide a related situation grounded in real-world settings for each cultural descriptor (e.g., travel consultation in Figure~\ref{fig:fig0}). % for each cultural descriptor, which facilitate the evaluation of LLMs' cultural awareness in practical settings. 
Then we evaluate state-of-the-art LLMs' cultural awareness on \dataname, and the results show room for improvement. Additionally, we demonstrate that training LLMs on \dataname enhances their performance on downstream culture-related tasks. We also show that our pipeline can be easily generalized to Reddit, another online community, illustrating its transferability and potential for future expansions.

To summarize, we make the following contributions. 
\begin{itemize}
    \item A general framework to collect cultural knowledge from online communities (\S\ref{sec:pipeline})
    \item \dataname, an open-source cultural knowledge base with 12K cultural descriptors from TikTok and 11K from Reddit (\S\ref{sec:dataset} and \S\ref{sec:generalizability}).
    \item Grounded evaluation on existing LLMs' cultural awareness (\S\ref{sec:main evaluation}) and a more culturally-aware language model fine-tuned on \dataname (\S\ref{sec:fine-tuning}) 
   %  \item future insights towards culturally-aware language technologies (\S\ref{sec:insights})
\end{itemize}
% First, we develop a generalizable pipeline to construct cultural knowledge bases from noisy data in online communities. Second, using the pipeline, we build and release \dataname, a knowledge base with 12K self-reported cultural descriptors from large online communities (TikTok and Reddit). Third, with \dataname, we evaluate existing LLMs' cultural awareness and propose a more grounded evaluation, showing that there is a big room for improvement. Next, we train a more culturally-aware language model on \dataname,  show that

\vspace{-1em}
\section{Related Work}
\vspace{-0.5em}
% previous work on cultural knowledge from social science 
%% e.g., https://ehrafworldcultures.yale.edu/
\noindent\textbf{Cultural knowledge bases.} 
% Cultural knowledge is the confluence of information accumulated through socialization and experiences and is an important dimension of social categorization ~\citep{Weinberger_2023}.
There have been many cultural knowledge base efforts in different domains \citep{lee2023crehate, kim2024click, jin2023kobbq, fung2024massively}. 
With traditional ethnographic methods, social scientists recorded cultural knowledge through existing historical accounts, ethnographic data, and cultural documents. For instance, behavioral scientists  compiled a collection of cultural materials, and released an online database named eHRAF (the Human Relations Area Files)~\citep{eHRAF}. In computer science studies, researchers employ computational methods to automatically construct datasets \citep{Smart} from large sources or curate data from crowd source workers \citep{lee2023crehate}. ~\cite{candle2023} built a pipeline to extract assertive cultural commonsense knowledge from C4 \citep{raffel2020exploring}, a large collection of Internet data, and \cite{fung2024massively} used Wikipedia and navigated to related online documents to extract cultural knowledge. Data from these sources are much cleaner compared to online communities, and often focus more on normative cultural indicators. Since culture is highly heterogeneous, we also need descriptive cultural expressions from sources like online communities. StereoKG ~\citep{deshpande2022stereokg} used Reddit and Twitter to extract cultural stereotypes for 5 religious groups and 5 nationalities, but due to the lack of proper filtering, the results are noisy.
% In this work, we utilize online communities where people are motivated to provide opinions about cultural practices, to complement the space. 
As an important complement to existing data sources, our work proposes a pipeline to process highly noisy online communities data on a large scale, and show that it can be easily generalized across different platforms, to provide valuable descriptive cultural knowledge.  %into the Human Relations Area Files (HRAF) since the 1930s and have recently made public the online database eHRAF ~\citep{eHRAF}. Wikipedia's Outline of Culture page also makes use of existing formal records of cultures and includes information on the arts, entertainment, and celebrations for each nation ~\citep{wikipedia}. 

\noindent\textbf{Cultural-awareness in language models.}
Previous works have studied cultural dimensions in language models \citep{gutierrez2016detecting, ramezani2023knowledge, jiang-etal-2020-x, adewole2021dialoguebased, yao2023empowering, li2024culturellm, Cao, liu2021visually, hämmerl2022multilingual, huang2023culturally, wang2023countries, köksal2023languageagnostic}. %Research on the cultural background of ChatGPT calls for greater diversity in cultural knowledge as English prompts biased model responses to American culture ~\citep{Cao}. 
On the evaluation side, prior studies have measured subjective global opinions from LLMs ~\citep{Durmus, santurkar2023whose}, and probed cultural value differences in these models ~\citep{arora2022probing, yin2022geomlama, roberts2023gpt4geo}.  %and experiments that test whether multilingual LMs capture different moral norms ~\citep{}. %There have also been studies that directly test cultural moral norms knowledge attainment in English language models and found that pre-trained English language models do not predict moral norms in different countries as well as previously reported ~\citep{ramezani2023knowledge}. 
On the model side, CultureLLM \citep{li2024culturellm} proposed a cost-effective method to integrate cultural differences into language models with augmented data. This work proposes a grounded way to evaluate cultural awareness to match real-world use cases, and fine-tune a more culturally aware language model with descriptive cultural behaviors constructed from online communities. 
% Other multilingual models can detect cross-cultural differences ~\citep{gutierrez-etal-2016-detecting} and have been tested on their abilities to reason in a multicultural context ~\citep{liu-etal-2021-visually}. In the realm of machine translation, research has been done to make LLM-based machine translations more culturally aware ~\citep{yao2023empowering}. Dialogue-based simulations also exist for cultural awareness training ~\citep{adewole2021dialoguebased}. 
% Another group of work looks into the cultural biases in language models, including studies on the cultural awareness of the Hindi language ~\citep{malik-etal-2022-socially}, cultural dominance ~\citep{wang2023countries}, cultural bias towards Western cultures ~\citep{naous2023having}, stereotypes ~\citep{cheng2023marked},  social biases  ~\citep{liang2021understanding}, nationality bias ~\citep{köksal2023languageagnostic}, and political biases ~\citep{feng2023pretraining}.
%Preventative measures have been taken to mitigate ethnic bias in BERT ~\citep{ahn2021mitigating}, but a systematic way to test the acquisition of both formal and informal cultural knowledge is needed. 

\vspace{-1em}
\section{\dataname Taxonomy}
Prior efforts on cultural knowledge base \citep{candle2023} often represent cultural knowledge in free-text sentences. But free-text contents on online communities are often noisy, and such an unstructured representation hinders further computational operation such as search and filter. Therefore, we develop a taxonomy (shown in Table~\ref{tab:taxonomy examples}) for more structured cultural knowledge representation, based on the taxonomy of social factors \citep{hovy-yang-2021-importance}, and the taxonomy of social norms \citep{ziems-etal-2023-normbank, goffman2002presentation}. It has the following fields: (1) \textbf{cultural group}, (2) \textbf{context}, (3) \textbf{goal}, (4) \textbf{actor}, (5) \textbf{recipient}, (6) \textbf{relation}, (7) \textbf{actor's behavior}, (8) \textbf{recipient's behavior}, (9) \textbf{other description}, (10) \textbf{topic}, and (11) \textbf{agreement}. For all these fields, we provide in-context examples and let the model extract any related information without constraint. This allows diversity and inclusivity in the data: for instance, examples for \textbf{cultural group} include typical cultural groups by countries such as ``\textit{American}'', as well as more fine-grained ones by regions or ethnicity groups such as ``\textit{Californian}'' and ``\textit{Asian American}'', and more broad social groups such as ``\textit{international students}'' which can be overlooked before \citep{barth2010introduction, stenou2002unesco}. %do not limit  the cultural group to geo-factors like countries and regions, and instead let the model extract any related cultural group mentioned in the text. So \textbf{examples} include typical cultural groups by countries such as ``\textit{American}'', more fine-grained ones by regions or ethnicity groups such as ``\textit{Californian}'' and ``\textit{Asian American}'', and more broad social groups such as ``\textit{people from small towns}'' \cite{barth2010introduction, stenou2002unesco}. For \textbf{context}, the model will also extract any related information without limitation, which allows us to capture cross-culture behaviors, such as ``\textit{Americans in France}'', while prior work most focused on intracultural behaviors. This field can be blank. 

\begin{table}[!tbh]
\centering
\resizebox{\columnwidth}{!}{
\begin{tabular}{p{3.2cm}p{6cm}p{7cm}}
\toprule
\textbf{Field}   & \textbf{Definition}             & \textbf{Example}                                                           \\
\midrule
Cultural group   & groups of people with similar cultural backgrounds    & American, Californian, Asian American, international student %people from small towns                 
\\
\midrule
Context    & settings the behavior takes place          & in France, in public, 4th of July celebrations                    \\
\midrule
Goal      & what the behavior aims to achieve           & to adapt to different cultures, to celebrate \\
\midrule

Actor     & who exhibit the behavior           &  people, customers, drivers                                                                \\
\midrule

Recipient  &  recipient of the action        & kids, service staff, passengers                                                                   \\
\midrule

Relation      & relation between the actor and the recipient       &  parents to children, actor to audience, among friends                                                                 \\
\midrule

Actor's behavior & behavior of the actor    &   dress casually, tip to express gratitude%,  drive on the right                                                         
\\
\midrule

Recipient's behavior & behavior of the recipient &  respond with thanks, 
accept card payments %, bag items for the customer                                                                 
\\
\midrule

Other description & anything that cannot fit into the other fields   &   Bangkok is known for its chaotic traffic                                                                \\
\midrule
Topic         & topic       &     education and technology, %consumer behavior, 
cultural exchange                                                              \\
\midrule

Agreement     & agreement level, \% of people who agree %\wyshi{clarify} 
&  an one-decimal float between 0 and 1, like 0.6                                                               \\\bottomrule
\end{tabular}
}
\caption{Fields, definitions and examples in the \dataname taxonomy. 
% \diyi{since you're creating this table to summarize these fields and examples, it might be nice to add a column called "definition", so that we do not need to repeat the definition so much in the text above}
}
\label{tab:taxonomy examples}
\end{table}

\vspace{-1em}
\section{Construction Pipeline}
\label{sec:pipeline}
\begin{figure}[ht]
\centering
\includegraphics[width=\textwidth]{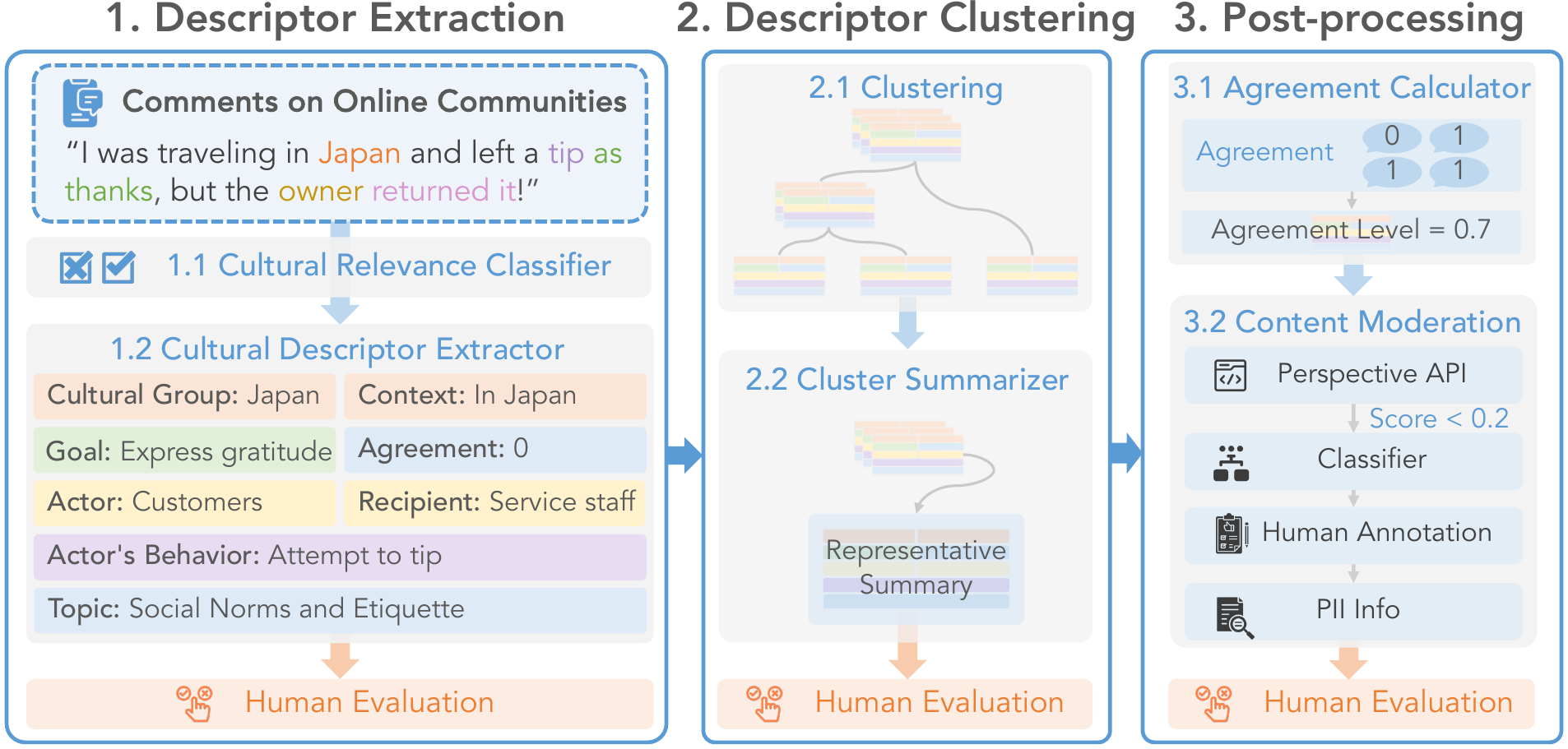}
\caption{\dataname construction pipeline. Starting from comments on online communities, we will (1) select culture-related comments and  extract mentioned cultural descriptors, then (2) cluster these descriptors and summarize the clusters, and finally (3) post-process them to get agreement value and remove bad contents. Each step is validated by human evaluation.}
\vspace{-1.5em}
\label{fig:construction_pipeline}
\end{figure}

Centering on the proposed taxonomy, we propose a bottom-up pipeline to construct cultural descriptors from online communities. Figure~\ref{fig:construction_pipeline} gives an overview of the pipeline which has three parts: (1) descriptor extraction, (2) descriptor clustering, and (3) descriptor post-processing.  See Section~\ref{appendix: pipeline} for more implementation details. 

\vspace{-1em}
\subsection{Descriptor extraction}
% Given the large amounts of noisy data
In the first part of the pipeline, we extract cultural descriptors from self-narrative data like comments and posts on online communities, and organize them into our taxonomy. 

\noindent\textbf{Culture relevance classifier.}
Given the large amounts of noisy data, the first step is to get the culturally-relevant portion. To do so, we annotate a subset with 280 training examples, and trained a distill-bert-based \citep{sanh2019distilbert} cultural relevance classifier. Then the classifier is applied on the entire dataset to get the subset related to culture. 
 The classifier achieves 
 an accuracy of 79\% on a held out test set with 100 examples.  
 % \caleb{Accuracy on a balanced test set?} 

\noindent\textbf{Cultural descriptor extractor.}
After obtaining the cultural subset, we employ Llama-2-70B \citep{touvron2023llama}, one of the best open-source LLMs at the task time, to extract values for each field in our taxonomy, by conditioning on the definition of fields and in-context examples. Listing~\ref{lst:knowledge_extraction_prompt} shows the prompt used. Human evaluation shows that this extractor 
% To validate the extractor's performance, we annotated a subset with 240 examples, and it 
achieves an accuracy of 82\% across fields in the taxonomy on a test set with 240 examples.

% Free-text contents on social media are often noisy, which hinders further operation , so the next step is to extract cultural knowledge and organize it into a taxonomy. 

\vspace{-1em}
\subsection{Descriptor clustering}
After the extraction step, we have many cultural descriptors, but the same cultural behavior can be expressed in many different ways, for instance, ``Japanese do not tip service staff'', or ``In Japan, people do not give tips''. So naturally in the second part, we need to first cluster the extracted descriptors, and summarize each cluster afterwards. %Moreover, since each individually extracted cultural descriptor may contain noise and errors, clustering also serves as a normalizer to ensure better data quality.

\noindent\textbf{Clustering.}
For the clustering step, we concatenate the extracted fields, encode the concatenated contents with SentenceBert \citep{reimers2019sentence}, and perform Hierarchical Agglomerative Clustering (HAC) clustering. %first concatenate the extracted fields and
% Then we conduct Hierarchical Agglomerative Clustering (HAC) on the cultural groups to identify similar cultural groups, and then enumerate each cultural group to perform another round of HAC inside each cultural group. 
We use the cluster size as the support value and remove clusters with less than 5 data points to ensure enough supporting evidences. 
The clustering parameters are chosen based on the performance on a validation set, and the clustered results achieve an average Silhouette score of 0.14 within the clusters. %We keep the year of the comments in each cluster, to help potential temporal analysis, as shown in Figure~\ref{fig:fig0}. 
 % discretized the cluster size into 10-unit intervals like $[10, 20)$ as the support value for each cultural descriptor. %The cluster size is  categorized into decadal intervals, and use it as the support number for each descriptors. 

\noindent\textbf{Cluster summarizer.}
After clustering, each cluster contains multiple cultural descriptors, so the next step is to summarize and generate a representative descriptor for each cluster. We use Mixtral-8X7B\citep{jiang2024mixtral}, a state-of-the-art open-source language model at the task time, to summarize each cluster. Since the clusters contain noisy opinions, the vanilla model often fails to output a comprehensive summarization with in-context examples. To achieve a better performance, we ask GPT-4 to generate 1K high-quality summarizations, and distilled those samples to train our own Mixtral summarizer.  % Using the model directly, however, produces suboptimal results even with in-context examples. As the clusters contains noise and different opinions, the vanilla model often fails to output a comprehensive summarization. 
Listing~\ref{lst:summarization_prompt} shows the prompt used for the summarizer. Human evaluation shows the cluster summarizer achieves a \textit{fidelity} score of 89.7\% and \textit{coherence} score of 96.6\%. Definitions of these metrics are available in \S\ref{appendix:clustering}.
% We evaluate the quality of cluster summarization results through human annotations, which reveal a \textit{fidelity} score of 89.7\% and \textit{coherence} score of 96.6\%. 

% Please refer to Section~\ref{appendix: pipeline} for more details such as topic normalization. 

 % \ryan{do we still want the "factualness" criterion?}

\vspace{-1em}
\subsection{Post-processing}
\label{pipeline:post-processing}

The final step is to post-process the clustered data. 

\noindent\textbf{Agreement calculator.} 
People may have different opinions regarding the same cultural behaviors, so instead of assertive statements, we provide agreement levels for each cultural descriptor in our \dataname. Each cluster now contains $\geq$ 5 data points, and each point is associated with an agreement score of 0 or 1, so we compute the average of these agreement scores as the agreement level. %As mentioned earlier, we also provide the 
Besides, the cluster size can also reflect the agreement level.

\noindent\textbf{Content moderation.} 
Finally, online platforms can contain controversial contents.  
So the last step is content moderation. To do so, we first use the perspective API \footnote{\url{https://perspectiveapi.com/}}, a machine-learning-based content moderation tool,  and filter out contents with scores above 0.2 for every category (toxicity, profanity, insult, identity attack, threat, severe toxicity). For more nuanced controversial contents, we annotate 800 examples and train a distill-bert-based classifier (test acc=0.77 on 117 examples), and employ a list of keywords to further identify them. Next, we manually label these identified contents, and remove bad ones. %359/2222these potential toxic or controversial content
Finally, we use the Presidio Analyzer\footnote{\url{https://microsoft.github.io/presidio/analyzer/}} to detect and remove Personal Identifiable Information (PII). %These steps removed around 1K cultural descri

\vspace{-1em}
\section{\dataname Dataset}
\label{sec:dataset}
\vspace{-1em}
TikTok is a popular social media platform with users from diverse cultural backgrounds, so we apply our pipeline on data from TikTok to construct our \dataname dataset. We obtain TikTok data via their official research API \footnote{\url{https://developers.tiktok.com/products/research-api/}} and collect a total of 34K %34431
posts and  
720K %726538
English comments from 2019/05 to 2023/08 with the hashtags ``\#culturaldifference'' and ``\#cultureshock''. %we follow the Research API Terms of Service to obtain the final \dataname dataset.  
Table~\ref{tab:basic stats} shows \dataname basic statistics after construction: for Tiktok, there are 12K cultural descriptors, 730 cultural groups, and 36 topics. Table~\ref{tab:Cultural Topics Distribution} shows the topic distribution. Table~\ref{tab:running time and number for tiktok} shows the running time and the data volume after each step.  % and notably, \yutong{259} cross-culture ones like ``American in France'', which are often missing in prior culture knowledge bases.   

 \begin{table}[htbp]
\centering
\begin{minipage}[t]{0.45\textwidth} % A minipage that covers half of the page width
\centering
% \resizebox{\textwidth}{!}{
\begin{tabular}{lcc}
\toprule
\textbf{Statistics} & \textbf{TikTok} & \textbf{Reddit} \\
\midrule
\# cultural descriptors & 11,754 & 11,236 \\
\midrule
\# cultural groups & 730 & 1,850 \\
\midrule
\# cultural topics & 36 & 36 \\
\bottomrule
\end{tabular}
% }
\caption{\dataname basic statistics.}
\label{tab:basic stats}
\end{minipage}
\hfill % Space between the minipages
\begin{minipage}[t]{0.45\textwidth} % Another minipage for the second table
\centering
\begin{tabular}{lcc}
\toprule
\textbf{Metrics} & \textbf{TikTok} & \textbf{Reddit} \\
\midrule
Well-formatted & 98.5\%&95.5\% \\
\midrule
Traceable & 93.3\%&94.0\% \\
\midrule
Meaningful & 84.5\% & 85.0\% \\
\bottomrule
\end{tabular}
\caption{Annotated \dataname quality. % quality annotated by humans
}
\label{tab:manual annotation}
\end{minipage}
\end{table}

To assess the dataset quality \textbf{quantitatively}, we select a random subset with 200 samples, and four human annotators annotated them for their (1) format (if the descriptor is well-formatted), (2) traceability (if it is possible to trace the cultural knowledge on the Internet) and (3) meaningfulness (if the descriptor provides meaningful cultural insights rather than generic ones). We evaluate them on traceability instead of factualness, since these descriptors are self-reported and may be nuanced, so it is difficult to fact-check them; as long as there is related information online, we consider it traceable and meaningful to be included.  The annotators achieved a Kappa score of 0.8. Table~\ref{tab:manual annotation} shows that  %We do notAs the annotators may not come from the cultural groups in the samples, they are instructed to search on the internet to see if the descriptor is factual or not. 
\dataname has well-formatted, traceable and meaningful cultural descriptors with moderate noise levels. %\wyshi{TODO: update, and add kappa later}

% \begin{itemize}
%     \item Formatting: Is the presentation and formatting of the information accurate and coherent?
%     \item Meaningfulness: Does the record provide significant insights into cultural behavior or information?
%     \item Factualness: Is the presented knowledge truthful and verifiable?
% \end{itemize}
% The summarization results are high-quality, achieving a pass rate of over 98\% in formatting, over 85\% in meaningfulness, and over 96\% in factualness. The inter-annotator agreement among three annotators is over 80\%.

Table~\ref{tab:data examples} shows \textbf{qualitative} examples in \dataname. It presents interesting features, such as: 
% \begin{itemize}
    % \item 
    (1) \textit{cross-culture behaviors}: e.g., Americans in France experience culture shock in terms of electricity bills and driving habits; 
    % \item 
    (2) \textit{linguistics variations}: e.g., Americans use ``chickpeas'' or ``garbanzo beans'' interchangeably; 
    % \item 
    (3) \textit{diverse ethnic groups}: e.g., Italian Americans identify themselves as Italian American with varying connection to Italy heritage; 
    % \item 
    (4) \textit{recent cultural information}: e.g., Chinese people heavily rely on mobile payment;
    % \item 
    and (5) \textit{cultural nuances} hard to obtain from formal sources like Wikipedia: e.g., in South Africa, some people express frustration over having to calculate prices and taxes separately while others do not think so.
% \end{itemize}

For the following evaluation (\S\ref{sec:main evaluation}) and fine-tuning (\S\ref{sec:fine-tuning}) steps, we split \dataname-TikTok by cultural descriptors into 9402 train, 1183 validation, and 1169 test samples.

\section{Evaluating LLMs' Culture Awareness}
\label{sec:main evaluation}
With \dataname-TikTok, we evaluate LLMs' cultural awareness. Prior work %casts cultural knowledge into a classification problem and 
 asks LLMs to answer cultural true/false questions \citep{fung2024massively}. But LLMs are used in contextualized  settings like a dialogue agent. So we propose  a grounded evaluation, that grounds cultural knowledge in a real-world scenario, %This not only evaluates their understanding of the cultural knowledge, but also 
 to test LLMs' ability to integrate cultural knowledge into their responses. %to see if the LLM can apply the knowledge correctly. 
 We also perform classification-based direct evaluation in \S~\ref{appendix:direct eval}. % two evaluation methods: (1) direction evaluation (\S~\ref{sec:direct eval}), that tests if the LLM can predict whether a certain cultural behavior is common, (2) grounded evaluation (\S~\ref{sec:grounded eval}), that grounds the cultural knowledge in a relevant scenario and checks if the LLM can apply the knowledge correctly. 

% \label{sec:grounded eval}

% \begin{figure}[ht]
% \centering
% \includegraphics[scale=0.4]{img/example_questions.png}
% \caption{Examples of the generated persona and questions. }
% \label{fig:fig_question_examples}
% \end{figure}

% To address the shortcomings in direct evaluation, we propose a contextualized evaluation method grounded in real-world application, which evaluates not only LLMs' understanding of the cultural knowledge, but also their ability to integrate the cultural knowledge into their responses in conversational consulting scenarios. 
% \raya{is it possible to make figure 4 a bit larger (rescaling leaves the table blurry)?}

\begin{figure}[htbp!]
\centering
\includegraphics[scale=0.43]{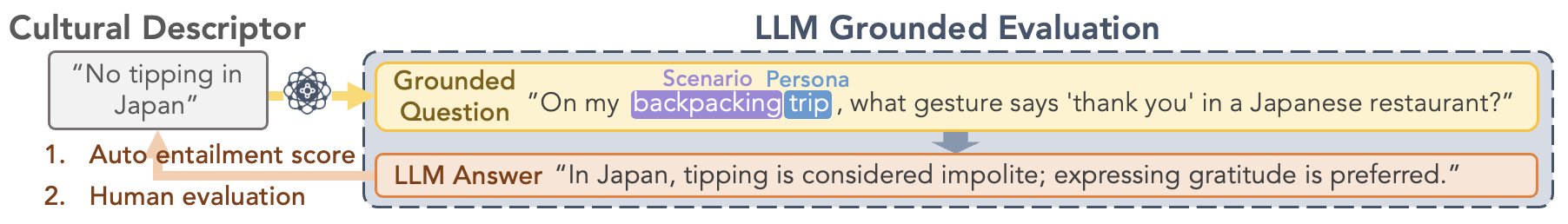}
\caption{Workflow of grounded evaluation. We present the grounded question to an LLM and get an answer. Given the answer, we perform \textit{automatic evaluation} and \textit{human evaluation}. 
}
\label{fig:grounded evaluation simple}
\end{figure}

\noindent\textbf{Grounded data generation.} %To generate grounded scenarios and questions, w
For each descriptor in \dataname, we first use a Mixtral-8x7B model fine-tuned on GPT-4-generated examples to generate a relevant consulting scenario, a client persona, and a grounded evaluation question. Then we employ a self-refinement method to improve the model generation based on two quality-control metrics at inference time. Figure~\ref{fig:grounded evaluation simple} shows a generated example (For the "No tipping in Japan" descriptor, the grounded question is "what gesture says 'thanks you' in Japan?"). Human annotation shows 86\% questions are grounded on the original descriptor. See \S\ref{appendix:grounded evaluation} for more details.

\noindent\textbf{Grounded evaluation.} As shown in Figure~\ref{fig:grounded evaluation simple}, %we first use LLMs to generate questions grounded on each cultural descriptor (more detail in \S\ref{appendix:grounded evaluation}), such as ``what gesture says 'thanks you' in Japan?'' 
% Then 
we present the generated grounded question to the LLM for an answer. Given the answer, we perform (1) automatic evaluation that uses GPT-4 to judge if the answer entails the original cultural descriptor (entailment score); and (2) human evaluation where two experts compare answers from two LLMs and select the more culturally-aware one (win rate). 

%We ground on 1K cultural descriptors, and ask GPT-4 to generate grounded scenarios, personas and questions \diyi{is the robustness check for this step done? }. Then we fine-tune a Mixtral model on these 1K examples, use a reward model to refine the generation, and finally use it to generate data for the entire set. Table~\ref{tab:grounded example1} shows an generated example. \S\ref{appendix:grounded evaluation} has more detail on the process. % for more detail.  %use a Mixtral model to generate related scenario, persona, and question, and ask it to self-refine the generation .  %such as travel condition on the cultural  generate related scenario
 %: the cultural descriptor is about parents leaving babies outside for naps in cold weather, %and the persona is someone from a warm climate moving to Reykjavik, 
%and the grounded question is a situation where the person is concerned about their colleagues' babies sleeping outside. 

\noindent % Ensures there is no indentation at the start of the minipage environments
\begin{minipage}[c]{0.48\textwidth}
\centering
% Replace 'table' with your actual table content
\resizebox{\textwidth}{!}{\begin{tabular}{lcccc}
\toprule
\textbf{Model} & \multicolumn{1}{c}{\textbf{High}} & \multicolumn{1}{c}{\textbf{Mid}} & \multicolumn{1}{c}{\textbf{Low}} & \multicolumn{1}{c}{\textbf{All}} \\
% \cmidrule{2-5}
% & Avg & \% Entail & Avg & \% Entail & Avg & \% Entail \\
\midrule
\midrule
Llama-2-7B-chat  & 71.2 & 66.0 & 61.2 & 62.5\\
\midrule
Llama-2-70B-chat  & 74.9 & 66.2 & 64.2 & 65.1 \\
\midrule
Mistral-7B-Instruct  & 72.9 & 67.2 & 63.4 & 64.5\\
\midrule
Mixtral-8x7B-Instruct  & 73.9 & 67.4 & \textbf{66.3} & \textbf{66.9}\\
\midrule
GPT-3.5  & 71.4 & 66.4 & 61.8 & 62.6\\
\midrule
GPT-4  & \textbf{75.8} & \textbf{67.9} & 65.0 & 66.1\\
\midrule
\midrule
\textbf{Llama2-7B-SFT (Ours)}  & \textbf{75.7} & 67.1 & 63.8 & 64.7\\
\midrule
 \textbf{Mixtral-8X7B-SFT (Ours)}  & 73.3 & 70.3 & 66.6 & 67.5 \\
\midrule

\textbf{Mixtral-8x7B-DPO (Ours)}   & 72.4 & \textbf{70.5} & \textbf{68.1} & \textbf{68.7}\\
% \midrule

% Mixtral-Augmented & 93.7 & 94.3 & 93.3 & 94.9 & 92.2 & 92.7 \\
\bottomrule
\end{tabular}
}
\captionof{table}{\label{tab: grounded eval by support} Automatic evaluation on LLMs' cultural awareness, evaluated by knowledge \textbf{entailment scores} on our grounded evaluation benchmark by support. \textbf{High support}: cluster size $> 50$ (70 examples). \textbf{Mid}: cluster size between 20 and 50 (175 examples). \textbf{Low}: cluster size $\leq 20$ (924 examples). }
\end{minipage}%
\hfill % Adds horizontal space between the minipages
\begin{minipage}[c]{0.48\textwidth}
\centering
% Replace 'example-image' with your actual image file name
\includegraphics[width=\textwidth]{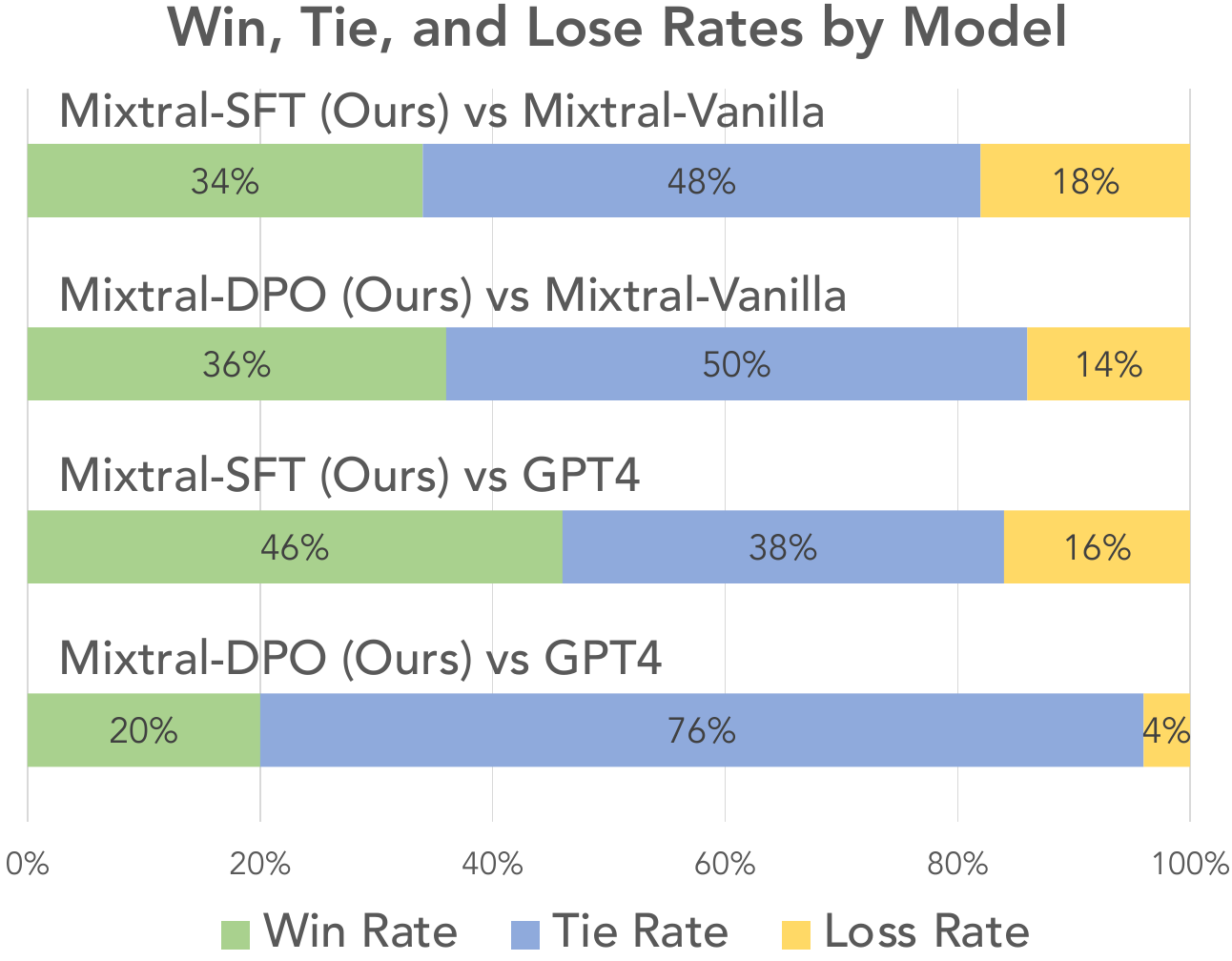}
\captionof{figure}{Human evaluation on \textbf{win rates} between different LLMs (50 examples per pair) evaluated by humans on cultural-awareness in grounded consulting scenarios. The two annotators achieved a Kappa score of 0.87.
\label{fig:win_rates}
}
\end{minipage}

\noindent\textbf{Evaluated models.} We evaluate open-source (Llama-2, Mixtral), close-source (GPT families \citep{achiam2023gpt}), and our own fine-tuned models (See~\S\ref{sec:fine-tuning}). See Table~\ref{tab:map between name and model cards} for the model version details. All the results are on the test set.  

% For evaluation, we ask LLMs to answer the generated grounded questions \diyi{i know we're short on space, but it seems a bit hard to understand this step without any example --- like what types of grounded questions? readers have to imagine what this looks like...recommend adding an example}, and use GPT-4 to judge if their answer entails the original cultural descriptor. %GPT-4's predicted probabilities of the answer being "Yes" is used as the entailment score. 
% Higher entailment scores mean better cultural awareness. %The detailed prompt template for knowledge entailment is in Listing~\ref{lst:knowledge_entailment_prompt}. 

\noindent\textbf{Automatic entailment results.} 
Table~\ref{tab: grounded eval by support} shows the average entailment score of each model split by the cluster size (level of support). %``mixtral-knowledge-augmentation'' shows the upper bound performance, where the LLM conditions on the gold cultural descriptor in the input to generate a culturally-aware answer. 
Mixtral-8X7B and GPT-4 are the best but still has a relatively low overall score of 66.9 and 66.1, suggesting room for improvements. Larger models have slightly better performance than their smaller versions. For more long-tailed cultural descriptors with fewer supports, the  performance drops as expected.  %We see that LLMs often make mistakes when XXX. 

% \yutong{win rate}
% \ryan{TODO:win rate+ qualitative analysis on GPT-4 performance + example in appendix}

\noindent\textbf{Human-evaluated win rate.} Figure~\ref{fig:win_rates} shows the human evaluation results. Compared to the base Mixtral model, Mixtral-DPO is strictly more culturally aware \textbf{36\%} of the time, and equally good \textbf{49\%} of the time; compared to GPT-4, Mixtral-SFT wins \textbf{46\%} of the time and ties \textbf{38\%} of the time. This trend also aligns with the automatic evaluation in Table~\ref{tab: grounded eval by support}, indicating that the automatic entailment evaluation makes sense. Qualitatively (Table~\ref{tab:qualitative_results}), we find that our fine-tuned models generate shorter, and more culturally specific answers tailored to the user's inquiry (e.g., ``in France, you might seek out artisanal cheese shops''), whereas standard RLHF-ed models like GPT-4 often give generic templated answers (e.g., ``1. Research local specialties, 2. Portion control...'').
% In addition to the automated knowledge entailment metrics, we also conducted double-blind expert evaluations to compare the cultural awareness and specificity of the model responses. According to human annotators,
% our best finetuned model performs equally as good or better than GPT-4 \textbf{84\%} of the times, and strictly better than GPT-4 in terms of cultural awareness \textbf{46\%} of the times. In comparison to the base Mixtral 8x7B model, our finetuned models wins \textbf{16\%} (SFT) and \textbf{36\% (DPO)} of the times, as ties \textbf{70\% (SFT)} and \textbf{50\% (DPO)} of the times. Our human evaluations also resulted in some interesting qualitative insights:

\S\ref{appendix: win rates} has more details on the human evaluation, models' win rates, and qualitative analysis.

\vspace{-1em}
\section{Fine-tuning a More Culturally Aware Language Model}
\label{sec:fine-tuning}
Our ultimate goal is to develop more culturally-aware language technologies. So we train on our \dataname dataset to see if such a resource can improve LLMs' cultural awareness. 

\noindent\textbf{Training process.} The training has two steps. First, we train a model on the 9402 cultural descriptors in the training set via supervised fine-tuning (SFT). In the second step, we select a 2K subset where the model performs poorly, and train on the grounded questions and answers augmented by golden cultural descriptors via SFT (``\textbf{model-SFT (Ours)}'' in the tables) or DPO \citep{rafailov2024direct} (``\textbf{model-DPO (Ours)}'') . See \S~\ref{appendix: fine-tuning} for more details.   %We split \dataname by cultural descriptors into 9779 train, \ryan{TODO: update numbers}1221 validation, and 1221 test samples. To fine-tune a culturally-aware language model, we first train the model on the 9779 cultural descriptors in the training set through supervised fine-tuning.  Then, we dynamically \wyshi{?} sample 2000 question where the model performance is unsatisfactory, use knowledge augmentation on these questions to generate more culturally-aware responses (Listing~\ref{lst:indirect_eval_aug_prompt}), and then perform either supervised-finetuning \wyshi{which one exactly} or direct preference optimization \citep{rafailov2024direct} on these 2000 examples.

\noindent\textbf{Results.} Table~\ref{tab: grounded eval by support} shows the results on the test set. The test set mostly contains out-of-domain cultural descriptors, but the fine-tuned models can still  achieve a better performance than their vanilla versions, suggesting that \dataname can improve models' cultural awareness. %These results suggest that \wyshi{hard to follow} training on a small set of culturally-aware QA samples can elicit the model's existing latent cultural knowledge and encourage the model to apply these cultural insights into its responses in grounded conversational scenarios. The models are able to generalize to cultural signatures that are unseen in the training dataset. 

\subsection{Zero-shot transferability on downstream tasks}
Ideally, a more culturally-aware language model can achieve a better performance on different culture-related tasks. So, we also evaluate the fine-tuned models on two downstream tasks, to see if our \dataname can help other cultural tasks in a zero-shot fashion. 

\noindent\textbf{Downstream tasks.} We choose two tasks: (1) GlobalOpinionQA \citep{Durmus}, which has questions from world value survey \citep{inglehart2000world} to measure how similar LLM-generated responses are towards different countries; (2) CultureNLI \citep{huang2023culturally}, which contains
premise-hypothesis pairs labeled by two cultural groups,  American and Indian. The detailed prompts and evaluation settings are in Appendix~\ref{sec:downstream_applications}.

\noindent\textbf{Results}. Table~\ref{tab: downstream tasks} shows the results. %For Global Opinion QA, the score shows the average similarity between the models' output and the surveyed distribution from each country; for Culture NLI, we show the prediction accuracy. 
On both datasets, the models fine-tuned on \dataname achieves a better performance than the vanilla counterparts (e.g., in GlobalOpinionQA, 79.5 VS. 81.8 for Mixtral-SFT; in Culture NLI, 59.9 VS. 61.5 in the US and 60.8 VS. 61.3 in India for Mixtral-SFT).
These results suggest that \dataname can be used to improve models' cultural awareness in downstream tasks even in a zero-shot setting.  

\begin{table}[h]
\centering
\begin{tabular}{ lccccc}
 \toprule
\multirow{2}{*}{\textbf{Model (zero-shot)}} &  \multicolumn{2}{c}{\textbf{GlobalOpinionQA}} & \hspace{0.5pt} &  \multicolumn{2}{c}{\textbf{CultureNLI}} \\
% \midrule
\cmidrule{2-3} \cmidrule{5-6}
 & \textbf{Avg Sim ($\uparrow$)} & \textbf{Skew ($\downarrow$)}&& \textbf{US ($\uparrow$)} & \textbf{IN ($\uparrow$)} \\
 \midrule
 \midrule
 Llama-2-7B-chat & \textbf{83.6}&\textbf{2.2}&& 39.2 & 39.5 \\
 \midrule
 % \midrule 
 Llama-2-70B-chat & \textbf{83.6}&\textbf{2.2}&& 69.7 & 68.9 \\
 \midrule
 Mistral-7B-Instruct & 79.3&3.2&& 42.5 & 43.8\\
 \midrule
 Mixtral-8x7B-Instruct &79.5&2.7&& 59.9 & 60.8\\
 \midrule
GPT-3.5 &-&-&& 75.0 & \textbf{73.0} \\
 \midrule
 GPT-4 & -&-&& \textbf{80.0} & 72.0 \\
 \midrule
 \midrule
 \textbf{Llama2-7B-SFT (Ours)} & \textbf{85.4} & \textbf{1.5}& & 39.2 & 39.6 \\
 \midrule
 \textbf{Mixtral-8X7B-SFT (Ours)} & 81.8 %$\uparrow$
 &2.8&& \textbf{61.5} & \textbf{61.3} \\
 \midrule
 \textbf{Mixtral-8x7B-DPO (Ours)} & 80.5&2.6&& 56.3 & 55.4 \\
 \bottomrule
\end{tabular}
\caption{\label{tab: downstream tasks} Zero-shot cultural awareness on GlobalOpinionQA and CultureNLI. A higher \textbf{Avg Similarity} means the model's output distribution is closer to the surveyed distribution for each country. A lower \textbf{Skewness} indicates the model's predictions are more balanced across countries (less variance). \textbf{US} and \textbf{IN} show the F1 score on US and India. In GlobalOpinionQA, GPTs' results are NA because we do not have access to their logit distributions. %This shows fine-tuning on \dataname improves the model's performance on downstream cultural tasks than their vanilla versions in a zero-shot fashion. 
}
\end{table}

\vspace{-1em}
\section{Generalizing to Reddit}
\label{sec:generalizability}
% \diyi{i appreciate this generalization perspective. however, from a fresh reader perspective, this part comes a bit unconvincing and preliminary to me. i worry that this won't add much value to us, but will cause reviewers to criticize. so i recommend removing this for now and add it to a longer version once we have more results in the next month.}
There are different online communities, so it is important to test if our pipeline can be transferred to other  platforms. So we apply our pipeline on Reddit, another online community, with the following customization. Table~\ref{tab:running time and number for reddit} shows the running time and data volume.

% To apply our pipeline on Reddit, we make the following customization to fit the data. 
\begin{itemize}
    \item \textbf{Culture Relevance Classifier}: similar to TikTok, we first search for tags like "\#culturaldifference" on Reddit, but tags are not used frequently on Reddit, so we also directly search for cultural keywords on both submissions and comments to identify cultural contents. See detail in \S~\ref{appendix:generalization on reddits}.  %because Reddit has much more data than TikTok, we first perform keyword search. %Considering Reddit has more personalized and user-defined tags, we extend to more tags like such as "\#cultureadaptaion" %"\#culturaldifference", '\#culturedifference', '\#cultureshock', '\#culturalshock', 
    % and "\#culturalexchange". %Additionally, 
    % Given the infrequent use of tags on Reddit, we further conduct keyword searches on both submissions and comments to identify culturally relevant contents.  
    As Reddit comments are much longer than TikTok comments, the TikTok-based cultural relevance classifier does not work well. %So we perform simple text search and classification. %We first source discussions with tags related to cultural differences. 
    % Considering Reddit has more personalized and user-defined tags, we extend to more tags like such as "\#cultureadaptaion" %"\#culturaldifference", '\#culturedifference', '\#cultureshock', '\#culturalshock', 
    % and "\#culturalexchange". Additionally, given the infrequent use of tags on Reddit, we further conduct keyword searches on both submissions and comments to identify culturally relevant contents. 
    So we %use GPT-4 to 
    annotate 1K examples, and train a new cultural relevance classifier for Reddit to get the culturally-relevant portion from the keyword-curated subset. Finally, we obtain 2.6M cultural comments. Considering the computation cost, we take a random subset of 528K cultural comments for the following processing steps. % from 2005/12 to 2022/1.  % we obtain 2.6M culturally related 
    % data with 1,249,504 submissions and 4,318,924 comments from 2005/12 to 2022/12.
    
    % \yutong{I think we implement the same method to train the classifier on both data, we just used different preprocessing methods on Reddit, and I added some detail above.} \yutong{Should we discuss the final selected dataset here or list in the following table of statistical results like TikTok dataset.}

    \item \textbf{Descriptor Extractor}: to achieve a better performance, instead of using few-shot Llama-2 extractor, we fine-tune a Mixtral-based extractor on 1K GPT-4-generated extraction examples to extract structured cultural descriptors from Reddit comments. 
    % \ryan{TODO: did we also distill the data?}
    % \ryan{TODO: did we do anything special for the cluster summarizer?}
\end{itemize}

%  

% Due to computation constraints, we processed 528K comments from 2005/12 to 2022/12, leading to 11K cultural descriptors. 
Table~\ref{tab:basic stats} shows the basic statistics: \dataname-Reddit contains 11K cultural descriptors and 2K cultural groups. Human annotation in Table~\ref{tab:manual annotation}  shows that it also contains high-quality data, suggesting that our pipeline can be easily generalized to a different platform. 
 % \yutong{Do we need to list all the keywords we used for the search?}

% \yutong{Cultural patterns may persist for centuries, but some patterns of culture may also transformed when mere decades \cite{putnam2000bowling
% } \cite{nisbett2001culture} \cite{hamamura2012cultures}. Social phenomena are one of the factors that may shift cultural patterns. In recent years, the COVID-19 pandemic has emerged as a formidable event impacting cultural groups worldwide. Our data collection from TikTok includes 24 behaviors related to COVID-19, and we could observe some interesting cultural transformations from these behaviors. For instance, people say that in Germany, people, and shop owners prefer cash over digital payments, especially in smaller shops and restaurants. However, people report that during the COVID-19 pandemic, people in Germany increased their use of cashless payments, which indicates a temporary transformation in cultural behaviors induced by the pandemic's circumstances. As we navigate to the post-pandemic period, it remains uncertain whether these behavioral shifts will permanently change Germany's payment practices. This observation emphasizes the transient nature of cultural patterns and highlights the importance of considering temporality as a significant factor in studies of cultural awareness. }

\section{Recommendation for Culturally aware Language Technologies}
\label{sec:insights}
\vspace{-1em}
Informed by results on \dataname construction  and analysis,  
% (\S\ref{sec:pipeline}, \S\ref{sec:dataset}), 
cultural awareness evaluation, %(\S\ref{sec:main evaluation}), 
and fine-tuning,
% (\S\ref{sec:fine-tuning}), 
we outline insights towards future culturally-aware language technologies. %, on data source, data content, data analysis, evaluation. % explore potential paths for future dataset development in the sections that follow.

% From constructing cultural knowledge bases \dataname and analyzing the developed \dataname, we show important opportunities and challenges towards future culturally-aware language technologies. 
% From analyzing current data landscape (\S \ref{sec:analysis}) and evaluating LLMs' performance (\S \ref{sec:model}), we unveil the most challenging aspects of social intelligence that remain unaddressed by existing data resources or model capabilities. Guided by insights from our results, we discuss possible future directions for dataset development below.

% As demonstrated by our work, the emergence of LLMs opens up many opportunities for cultural knowledge base construction and culturally-aware language technologies.
\subsection{Cultural knowledge data}
We show that fine-tuning on \dataname can improve the cultural-awareness on various downstream tasks, so it remain critical to keep developing cultural knowledge databases. 

\noindent\textbf{Data source.} Prior work often relies on formal data sources and collapse different sources together: e.g., \cite{fung2024massively} started from WikiPedia and continued to scrape any related websites to construct cultural knowledge bases. 
% \cite{candle2023} extracted cultural knowledge from C4 \citep{raffel2020exploring}, a big Internet data dump with many data sources. 
But different data sources %host different populations and 
cover various aspects of culture: official documents like textbooks provide factual cultural knowledge, while online communities like social media offer insights on everyday cultural practices. So we should invite diverse data sources to capture the full spectrum of culture in the future. Besides, different data sources host different populations: Table~\ref{tab:Cultural Topics Distribution} shows that topic-wise, \dataname-Reddit contains more contents on community and identify, while \dataname-TikTok is more about daily life like social norms and etiquette. So future datasets should keep data source as an important attribute to allow further analysis. %In this work, we source from two social media platforms, TikTok and Reddit, and keep the data source information available for further analysis.%\cite{fung2024massively} started from Wikipedia and continued scraping related websites. 
% Besides, %
% official documents like textbooks can provide factual cultural knowledge, while online communities like social media offer insights on everyday cultural practices.  %Besides, different data sources host different populations' opinions,  and this leads to various aspects of cultural knowledge. Therefore, we should also keep data source as an important attribute in the final dataset. In this work, we source from two social media platforms, TikTok and Reddit, and keep the data source information available for further analysis. %TODO: We found that XXX
% official, self-reported anacadotes. 
% factual knowledge, or cultural norms

% we should consider various platforms, as culture is highly dynamic, and different platforms have different user population, rather than collapsing different sources together (wikipedia, C4, etc). 

% \paragraph{Data content.} Culture is highly dynamic and here is a list of dimensions to consider
\paragraph{Data contents.} Culture is multifaceted, so it is also important to factor in various dimensions in the data content. Here is an example list of attributes to consider. 

\begin{itemize}
    
    \item \textbf{Cross-culture behavior}. %Current datasets mostly contain the cultural practice of dominant groups within their specific region like Americans in the US. 
    In a globalized world, it is crucial to understand cross-culture behaviors to facilitate effective communication \citep{watkins2012learning}. Our \dataname contains some cross-culture behaviors but we need more efforts on it. 
    \item \textbf{Perspectives}. It is also important to track through whose lens we are looking at a certain culture behavior, because different perspectives may lead to different understanding of the same cultural practice \citep{iyengar1999independence, brewer1999psychology}. %It is important to track through whose lens we are looking at a certain culture. %For instance, for the tipping culture,  when an American
% show something
% \item \textbf{Agreement level}. Cultural practice varies from people to people, in our work we release the agreement level towards different cultural practices, which can help more flexible understanding of the complex nature of Culture. 
\item \textbf{Time}. Culture changes over time. In \dataname, we release the time range associated with the cultural descriptors. Future data efforts should also consider the time factor to enable temporal analysis. %In this work, we perform some preliminary analysis on temporal change in culture and release the years associated with the cultural descriptors. For instance, from \dataname we found that before COVID, German prefer cash over digital payments; but during the COVID-19 pandemic, people in Germany increased their use of cashless payments. This indicates a temporary transformation in cultural behaviors induced by the pandemic's circumstances. This observation emphasizes the transient nature of cultural patterns and highlights the importance of considering temporality as a significant factor in studies of cultural awareness. But existing work often overlooks the time factor and future cultural data efforts should also consider the temporal factor in culture to understand how culture develops over time. 

    \item \textbf{Multilingual}. Culture and language are deeply intertwined. But many existing cultural knowledge bases still rely on English. To capture the cultural nuances, in the future, we should develop multilingual multicultural knowledge banks.

% In recent years, the COVID-19 pandemic has emerged as a formidable event impacting cultural groups worldwide. Our data collection from TikTok includes 24 behaviors related to COVID-19, and we could observe some interesting cultural transformations from these behaviors. For instance, people say that in Germany, people, and shop owners prefer cash over digital payments, especially in smaller shops and restaurants. However, people report that during the COVID-19 pandemic, people in Germany increased their use of cashless payments, which indicates a temporary transformation in cultural behaviors induced by the pandemic's circumstances. As we navigate to the post-pandemic period, it remains uncertain whether these behavioral shifts will permanently change Germany's payment practices. This observation emphasizes the transient nature of cultural patterns and highlights the importance of considering temporality as a significant factor in studies of cultural awareness. 

\item \textbf{Multimodality}. Cultural knowledge goes much beyond text information.  So it is essential to include different modalities to capture the full spectrum of culture, from non-verbal communication cues, to rituals and arts, and so on in the future.  
\end{itemize}

\paragraph{Data analysis.} In terms of data analysis, future research should consider \textbf{temporal change} rather than focusing on static data, as culture is evolving over time. For instance,  we perform preliminary temporal analysis in \S~\ref{appendix:temporal analysis} and find there are more discussions around studying abroad, LGBTQ+ rights, and technology over the years. Besides, 
 existing research still categorizes culture by country, but we need to attend to more \textbf{fine-grained cultural groups} (ethnicity, generation, regions, ethnolinguistics, immigrants, socioeconomics, etc), to fully understand cultural diversity. Moreover, the study of \textbf{cultural adaptation} becomes increasingly important, as it reveals how culture changes in response to global influences. These focus areas -- temporal dynamics, cultural group diversity, and adaptation processes -- offers a comprehensive understanding of the fluid nature of culture in a globalized world.

% \subsection{data analysis}
% \paragraph{temporal}
% \paragraph{spatila ethnicity group} break the boundry between countries. 
% \paragraph{cultural adaptaion}
% \paragraph{contextualized}

\subsection{Cultural awareness evaluation}
% We also show that different evaluations could provide different insights on cultural awareness. 
% In particular
We highlight two findings in evaluation. First, in our evaluation, humans also find it difficult to decide which model response is more culturally aware, partly because they are not from the presented cultural group. As we spend more effort on cultural data resources, it is also increasingly important to involve \textbf{global annotators} to enable more accurate evaluation. Secondly, as shown in our findings, direct and \textbf{grounded evaluations} give different results. %, and direction evaluation is different from the actual real-world use cases.  %And current evaluation methods rely so much on direction evaluation, which is far away from the actual real-world use cases. 
So during evaluation, it is important to be more grounded on the end applications.

\subsection{Training culturally-aware language technologies}
% zero-shot transfer? is DPO or SFT better
% \ryan{DPO is generally better when we have limited finetuning data, but SFT is more stable and faster to train. I don't think we need to comment on DPO vs SFT in this paper though?}

We realize that when fine-tuning models for cultural awareness, training only on the cultural knowledge or the grounded QA tasks could be insufficient. Take training a culturally-aware conversational assistant as an example. First, it requires appropriate cultural data grounded in \textbf{multi-turn conversational} settings. In addition, it requires a \textbf{well-designed training paradigm} to attend to the cultural nuances potentially implicit in the dialogue context. It also needs a \textbf{solid evaluation method} to rate the culture awareness of the generated responses, to help the model improve and evolve. Such a model needs to have a holistic view of the user cultural background, a personalized recognition of individual differences, and an inclusive mind for new cultural concepts and practices.  %know more than individual cultural behaviors or signatures, but connect the behaviors together and understand that people with different cultural context might have different expectations and choose appropriate strategies to engage in the conversations.

% We can also mention that multi-turn conversations presents further complications, as we need to ensure that the model can not only provide culturally-appropriate responses in a single turn, 

% \ryan{We can highlight that when finetuning models for cultural awareness, training only on the cultural knowledge or through direct classification tasks could be insufficient. Training a culturally-aware conversational assistant is more involved, and requires careful finetuning on data grounded in conversational scenarios to elicit the model's ability to integrate cultural knowledge into conversational settings.}

% \ryan{We can also mention that multi-turn conversations presents further complications, as we need to ensure that the model can not only provide culturally-appropriate responses in single turn, but also consistently pay attention to the cultural nuances throughout the entire conversation. This requires the model to know more than individual cultural behaviors/signatures, but connect the behaviors together and understand that people with different cultural context might have different expectations and choose appropriate strategies to engage in the conversations.}

\section{Conclusion}
To conclude, our study introduces a generalizable pipeline for creating cultural knowledge bases from online communities. Using the pipeline, we develop \dataname, a cultural knowledge database with 12K cultural descriptors sourced from TikTok, and 11K from  Reddit. \dataname features agreement levels for nuanced cultural interpretation and contextualized scenarios for grounded evaluation. %, marking a significant advancement over existing cultural knowledge tools. 
With \dataname, we assess the cultural awareness of various LLMs, showing room for improvement. Further, fine-tuning an LLM with \dataname leads to better performance on downstream cultural tasks, which showcases  the potential of \dataname. Finally, drawing from our findings, we close the paper by presenting insights towards future culturally-aware language technologies.
\section*{Ethical Statement}
In this work, we construct a cultural knowledge base from online communities. %Although we have tried our best to address the complex ethical landscape that accompanies such work. 
Given the large size of the dataset, we acknowledge that stereotypes, controversial, and negative content may still exist in our dataset, despite our rigorous efforts to filter the data and minimize the impact of such content. We want to emphasize that the cultural descriptors in \dataname are not intended to reflect, nor should they be interpreted as reflecting, the personal views or opinions of the authors or the online platforms.
 We call for a better approach for content moderation in the future and hope that researchers will use our data with a discerning perspective, and always consider the broader implications of its application and the potential for reinforcing harmful biases. %. We implemented multiple layers of scrutiny, and aimed to minimize the presence and impact of such content. However, the dynamic and nuanced nature of cultural expressions means that complete eradication of problematic elements is an ongoing challenge.

We also recognize the responsibility that comes with handling cultural data, especially from diverse and broad communities like those on TikTok and Reddit. In our method, we have strived not only for technological innovation but also for a conscious approach that respects the dignity, privacy, and cultural sensitivities of individuals and groups represented in the data. This includes anonymizing data where possible, ensuring compliance with platform terms of service, and engaging with ethical guidelines that govern research in social sciences and humanities.

% Furthermore, we emphasize the importance of continual ethical reflection and dialogue within the research community. Our work is a step in a larger journey towards understanding and respecting the complex mosaic of global cultures. We advocate for the development of more sophisticated tools and methodologies that can discern and mitigate biases, stereotypes, and harmful content, enhancing the positive impact of language technologies on society.

% We also encourage researchers and practitioners using our dataset and findings to apply their ethical judgment, adapt their use cases to be sensitive to cultural contexts, and contribute to an ongoing process of improvement and refinement. Our aim is to foster a research environment that prioritizes ethical considerations, inclusivity, and respect for cultural diversity, setting a precedent for future endeavors in culturally-aware language technology development.

In conclusion, while we acknowledge the limitations and challenges inherent in our work, we believe in its potential to contribute positively to the field of culturally-aware language technology. We encourage the community to join us in these efforts, to promote cultural diversity, inclusivity, and sensitivity. We discuss limitations of this work in \S\ref{sec:limitation}.

% \section*{Author Contributions}
% % \thanks{
% Stanford processed the raw data internally. IBM provides high-level feedback and is not involved in the data processing. 

\section*{Acknowledgement}
We thank feedback from Chunchen Xu, Emily Goodwin, Jing Huang,  and members from the SALT lab at Stanford University. We also thank TikTok for providing the research API. Stanford processed the raw data internally. IBM provides high-level feedback and is not involved in the data processing.
% In summary, these limitations highlight the need for a more nuanced approach to the study of culturally-aware language technology. As mentionFuture research should aim to incorporate multilingual data, account for diverse perspectives within cultural analyses, provide detailed contextual information to interpret generic statements accurately, and address sample biases to present a more balanced and comprehensive view of cultural diversity.

% English-only data

% Perspective  

% generic statements

% limitation: sample/selection bias during data collection (e.g., many people only post on social media when they like or agree with a behavior, but not when they disagree), llama extraction bias, 

\bibliography{colm2024_conference}
\bibliographystyle{colm2024_conference}
\newpage
\appendix

\section{Limitations}
\label{sec:limitation}
Despite our efforts, we recognize several limitations of our work. 

First, in our pipeline, we utilize open-source LLMs in various steps. Constrained by open-source LLMs' ability to process non-English languages, we process English-only data. But many cultural nuances cannot be fully expressed or captured  by English. This limitation inherently restricts our ability to grasp and represent the full spectrum of cultural contexts and meanings, potentially leading to oversimplifications of certain cultural aspects. Besides, although we attempt to minimize bias with various efforts, these open-source LLMs could still extract biased information, and generate biased summarization, which could lead to biased final outcome in the data.

Second, our dataset is subject to sample bias. Because we scrape the data with certain keywords and hashtags like ``cultural difference'' and ``culture shock''. Oftentimes people only post on online platforms when they have strong reactions such as surprise or shock towards certain cultural phenomena. %  predominantly featuring expressions that evoke a strong reaction, such as shock or surprise. 
This bias means that our findings might overemphasize aspects of cultural difference that are more likely to  stand out to individuals, while underrepresenting more mundane or universally shared aspects of culture. Such a bias can skew the perception of cultural diversity and difference, potentially reinforcing stereotypes or overlooking the subtleties of cultural exchange and adaptation.

Third, \dataname still contains generic cultural statements, such as expressions of culture shock without detailed information, which may not help to provide nuanced understandings of intercultural interactions.

% lagging

\section{\dataname examples}
Table~\ref{tab:data examples} shows qualitative examples in our \dataname-TikTok. 
\begin{table}[H]
\centering
\resizebox{\columnwidth}{!}{
\begin{tabular}
% {p{3cm}|p{3cm}|p{2cm}%|p{1.5cm}
% |p{7cm}|p{5cm}|p{2cm}
% % |p{2cm}|p{4cm}|p{2cm}|p{1cm}|p{2cm}
% }
{p{3cm}p{3cm}p{2cm}%|p{1.5cm}
p{7cm}p{5cm}p{2cm}
% |p{2cm}|p{4cm}|p{2cm}|p{1cm}|p{2cm}
}
 \toprule
 \textbf{Cultural group} & \textbf{Context} 
 % & \textbf{Goal} & \textbf{Relation} 
 & \textbf{Actor} %& \textbf{Recipient} 
 & \textbf{Actor's behavior} 
 % & \textbf{Recipient's behavior} 
 & \textbf{Other description}
 % & \textbf{Topic} 
 & \textbf{Agreement} 
 % & \textbf{support num} 
 \\
 \midrule
 American&	in France		
 % &&
 &		people	
 % &
 &	experience culture shock, express surprise, and feel confused due to differences in lifestyle, food, and social norms 
 % &
 &		includes specific examples like electricity bills and driving habits 
 % &	cultural adaptation 
 & 0.9 
 % & [180, 190)
 \\\midrule

 American	& in the United States and grocery stores 
 % &	&
 &		people%&	
 &	refer to chickpeas as 'chickpeas' or 'garbanzo beans', often using the term interchangeably
 % &
 &		the term 'chickpeas' has been adopted from Hispanic language 
 % &	Cultural Traditions and Festivals 
 & 1.0
 % &[5,20) 
 \\
\midrule
 % American & in public & & & people & & dress casually, often in comfortable clothing, with a preference for sweatpants and following dress codes & &&Dress Codes& highly agree\\
 % \midrule 
 % Latin American &	Quinceanera celebrations &	mark a significant milestone in a girl's life&	celebratory and coming-of-age&	families and individuals	girls turning 15&&	celebrate a girl's 15th birthday with traditional rituals, including religious ceremonies, family gatherings, and gift-giving&&		costs can be high and the celebration can last for several days&	Cultural Traditions and Festivals & highly agree
 Italian American &	primarily in the United States	
 % &&
 &		individuals and communities	%&
 &	identify as Italian American, often with varying levels of connection to Italian heritage and culture	
 % &
 &	discussions around appropriateness and cultural preservation
 % &	Community and Identity 
 &	1.0	
 % &[5, 20)
 \\
 \midrule 
 Californians &	in California and its various regions	
 % &&
 &		people	%&	
 &experience a mix of high living costs, attraction to the state, and a preference for living there despite cheaper alternatives
 % &	
 &	California is perceived as wealthier, with varying expenses and a need for affordable housing	
 % & Miscellaneous 
 &	0.7
 % &	[20, 30)
 \\
 \midrule 
  Alabamian &	in Alabama and during road trips
 % &&
 &		people%& 
 &		enjoy outdoor activities and unique experiences like visiting In and Out and riding in the back of a truck 
 % &
 &
 % &			Entertainment and Leisure	
 &	1.0	
 % &[5, 20) 
 \\
 \midrule
 % \midrule 
Norwegian&	in Norway, particularly in the north 
% &	to avoid mistaking drugs for candies and maintain hygiene &	parent to child
&	people, including children and parents%&	children 
&	follow a strict candy consumption schedule, eating candy only on Saturdays and avoiding unwrapped candies	
% &
&	candy is considered a treat and is typically bought on Saturdays
% &	Food and Dining 
&	0.8
% &	[20, 30)
\\
\midrule
Chinese	&in China, particularly in urban areas 
% &	to make payments and transfer money&	customer to bank or store
&	people and businesses%&	banks and stores
&	heavily rely on digital and mobile payment methods like WeChat Pay and Alipay, often using facial recognition
% &	enable payments and receive payments
&
% &		Finance and Economy
& 1.0	
% &[20, 30)
\\
\midrule

Rwandan&	in Rwanda and among Rwandan communities
% &&
&			people%&
&		speak Kinyarwanda, Swahili, and English, with Kinyarwanda being the primary language
% &	
&
% &		Communication and Language
& 1.0	
% &[20, 30)
\\
\midrule
South African &	when paying for items	
% &&
&		people	%&
&	express frustration over having to calculate prices and taxes	
% &
&	prefer straightforward pricing without additional calculations	pricing expectations
% & Consumer Behavior
&	0.1	
% &[210, 220) 
\\
\midrule

Australian &	in Australia, particularly in restaurants and bars
% &	express gratitude&	customer to service staff
&	customers%&	service staff
&	tipping is not common or expected due to fair wages and good service
% &	receive tip
&	tipping is not a common practice in Australia, but can be seen in some high-end establishments
% &	Social Norms and Etiquette
&	0.5	
% & [50, 60) 
\\
\midrule
 
Argentinian&	in the northern regions including Jujuy, Salta, and the north
% &&
&			people%&	
&	enjoy spicy food, particularly in local cuisine
% &			
&
% &	Food and Dining  
& 0.7	
% &[5, 20)	
\\%\midrule
 \bottomrule
\end{tabular}
}
\caption{Selected qualitative examples in \dataname. We omit several fields for space. Please refer to the released dataset for the complete examples.} %\diyi{you can skip some columns to show these examples.. like no need to include "goal, relation, receipient" if they do not occur very often, and put the very detailed table to appendix}
\label{tab:data examples}
\end{table}

\section{Construction Pipeline}
\label{appendix: pipeline}

We present more details on the construction pipeline in this section. 
\subsection{Descriptor extraction} \label{sec:knowledge_extraction}
\noindent\textbf{Cultural descriptor extractor.} 
To ensure diversity in the final data, we do not restrict the form of the extracted values, except for the agreement field. For agreement, we ask the model to output 1 if the comment agrees with the extracted cultural information and 0 otherwise, to enable agreement calculation in \S\ref{pipeline:post-processing}. 

\paragraph{Negation converter} After extracting the cultural descriptor, we can also optionally perform a negation conversion step. This is because cultural descriptions can be diverse and dynamic, it is important to calculate the agreement levels to certain cultural behavior inside a cultural group. This requires both positive and negative examples for one cultural behavior. But due to the limitation of clustering algorithms, opposite sentences such as ``Japaneses do not give tips'' and ``Japaneses give tips'' may be placed into the different cluster in latter steps. So after the extraction and before clustering, we use different heuristics to convert negative sentences to their positive forms, and also flip the agreement field, i.e., ``Japanese do not give tips, agreement=1'' will become ``Japaneses give tips, agreement=0'', to help future clustering and agreement calculation steps. 

\subsection{Descriptor clustering}
\label{appendix:clustering}
\noindent\textbf{Clustering.} To manage the computation, we perform two rounds of clusterings. We first conduct Hierarchical Agglomerative Clustering (HAC) on the cultural groups to identify similar ones. Then we enumerate the identified cultural groups. Within each cultural group, we concatenate the extracted fields, encode the concatenated sentence with SentenceBert \citep{reimers2019sentence}, and perform another round of HAC clustering. %first concatenate the extracted fields and
% Then we conduct Hierarchical Agglomerative Clustering (HAC) on the cultural groups to identify similar cultural groups, and then enumerate each cultural group to perform another round of HAC inside each cultural group. 
% The clustering parameters are chosen based on the performance on a validation set (with an average Silhouette score of 0.14 within the clusters). 
We keep the year of the comments in each cluster, to help potential temporal analysis, so each descriptor is associated with a time range as shown in Figure~\ref{fig:fig0}. 
To ensure enough supporting evidence for each cluster, we remove clusters with less than 5 data points, and discretized the cluster size into 10-unit intervals like $[10, 20)$ as the support value for each cultural descriptor. %The cluster size is  categorized into decadal intervals, and use it as the support number for each descriptors. 

\noindent\textbf{Summarization.} To summarize each cluster into a single, high-level cultural indicator, we leverage a Mixtral 8x7B model finetuned on 1k summarization examples generated by GPT-4. To evaluate the performance of our summarizer, we performed human annotation on the following two metrics:
\begin{itemize}
    \item \textit{Fidelity}: extent to which the summary accurately represents the main ideas, facts, and figures from the original text without introducing inaccuracies or distortions; 
    \item \textit{Coherence}: How well the summary flows from one sentence to another, maintaining logical progression and clear connections between points.
\end{itemize}

\paragraph{Normalization on cultural group and topic.} 
Since we leverage LLMs to perform cluster summarization, and models can extract and summarize to generate any information, similar values may appear in different forms, for instance, ``Aboriginal Australians'' and ``Indigenous Australians'' are synonyms, %\caleb{Aboriginal people are a subset of the Indigenous peoples of Australia? A quick search says that other indigenous people are those of the islanders of the Torres Strait. Maybe specify that you collapse synonyms and hypernyms or something like that. The word `similar' could be misinterpreted.} 
and ``fashion and attire'' and ``clothing'' are two similar topics. Having such synonyms makes it hard to query and manage our dataset, so we normalize values in the two fields of ``cultural group'' and ``topic'', to collapse synonyms. Similar to the previous clustering step, we use HAC to group synonyms in cultural groups together and take the majority vote as the representative cultural group.
Grouping together similar topics turns out to be a more involved task, as many topics overlap with and differ from each other in subtle ways, so embedding-based clustering methods (e.g., HAC) often yield noisy results that fail to align with human judgements. Instead, we use an LLM to detect 100 recurring themes from our original topics, and then manually merge, prune, and rephrase these themes into 36 high-level cultural topics. Finally, we ask an LLM to classify each summarized cluster into one of these topics. Table~\ref{tab:Cultural Topics Distribution} shows the details of our final list of 36 cultural topics. 
% \ryan{cultural group normalization how is it done?}

\subsection{Post-processing}
\noindent\textbf{Agreement calculator.} We calculate the average of individual agreement scores and round it to the first decimal point as the final agreement level in our final \dataname data. 

\subsection{LLM Prompts for Different Steps}
Listing~\ref{lst:knowledge_extraction_prompt} shows the prompt used for the descriptor extraction step, and Listing~\ref{lst:summarization_prompt} shows the prompt used for the cluster summarizer. 
\begin{lstlisting}[language={}, caption=Prompt for Llama-2-70B on descriptor extraction, label=lst:knowledge_extraction_prompt]
[INST] <<SYS>>
You are a helpful, respectful and intelligent assistant trained to identify and extract cultural information. Your role is to follow the given instructions precisely and format your responses as required. Keep your responses succinct and limited to the requested information. If you don't know the answer to a question, please don't share false information.

Cultural information encompasses content that showcases the distinctive characteristics, artifacts, or manifestations of a specific group, community, or region. This includes, but is not limited to, practices, behaviors, norms, values, beliefs, habits, customs, architectural styles, environmental engagements, and any other elements that are emblematic of a particular cultural setting. It does not include generic information or widespread practices that are not distinctly tied to a specific cultural identity.

For this task, consider information as "cultural" if:

1. It is associated with or characteristic of a specific identified group (e.g., Americans, Italians, midwestern Americans, etc.).
2. It reveals a unique aspect of that group's way of life, including social conventions, physical creations, or interactions with their surroundings that are not typically seen in other cultures.
3. It provides insight into the cultural uniqueness, whether through social practices, material culture, or other culturally significant elements.

Please exclude generic or ubiquitous statements or observations that do not clearly relate to the unique cultural context of a specific group.
<</SYS>>
For each video-comment pair, you need to do two things:
1. Determine whether the provided example contains cultural information.
2. If the example does include cultural information, extract the cultural knowledge into a list of JSON objects with the following fields:
```
{
    "cultural group": "group of people with the same cultural background",
    "context": "location, or other settings this behavior is performed",
    "goal": "goal of the behavior",
    "relation": "relation between the actor and recipient",
    "actor": "the actor of the action",
    "recipient": "the recipient of the action",
    "actor's behavior": "the behavior of the actor",
    "recipient's behavior": "the behavior of the recipient",
    "other descriptions": "any other description that doesn't fit into previous categories",
    "topic": "cultural topic",
    "norm": "whether the described event is considered norm according to the given comment. 1 = norm; 0 = taboo",
}
```
If an example contains multiple cultural knowledge, please encode each piece of knowledge into a seperate JSON object.
Output the extracted cultural knowledge as a list of JSON objects, or an empty list if the provided example does not contain any cultural information.

----------------------------------------------------
Here are some examples: 
{few\_shot\_examples}

----------------------------------------------------
Now determine if the following example contains cultural information and extract any cultrual knowledge into a list of JSON objects. Please only include information that you directly extract from the provided text and do not hallucinate.

[Reminder]: Consider information as "cultural" if:
1. It pertains to a specific identified group (e.g., Americans, Italians).
2. It shows unique cultural traits or practices of that group differing from others.
3. It provides insight into the cultural uniqueness, whether through social practices, material culture, or other culturally significant elements.
Please avoid considering generic statements or behaviors that are common across multiple cultures or lack specificity as "cultural information."

Please base your answers strictly on the provided text. If important cultural context, such as the cultural group, is not explicitly mentioned or directly inferable from the text, output an empty list. Avoid adding or assuming any information that is not directly supported by the text.
Once you've outputed a list of JSON objects, please immediately output "<EOD>".

Video description: {}
Comment: {}
Contain cultural knowledge: [/INST]
\end{lstlisting}

\begin{lstlisting}[language={}, caption=Prompt for Mixtral-8X7B on cluster summarizer, label=lst:summarization_prompt]
<s>[INST] Here is a list of cultural behaviors that belong to a single cluster:
{}

Please summarize the above records into one consolidated JSON object. For each field in the output JSON, provide a concise, overarching summary that encapsulates the key essence or common theme of all records.

Your JSON output should contain the following fields:
{
    "cultural group": "group of people with the same cultural background",
    "context": "location, or other settings this behavior is performed",
    "actor": "the actor of the action",
    "recipient": "the recipient of the action",
    "relation": "relation between the actor and recipient",
    "actor's behavior": "the behavior of the actor",
    "goal": "goal of the actor's behavior",
    "recipient's behavior": "the behavior of the recipient",
    "other descriptions": "any other description that doesn't fit into previous categories",
    "topic": "cultural topic relating to the behaviors",
}

Your summary should:
1. Focus on the cultural aspect and ignore any non-cultural information
2. Leave a field as null if it is inapplicable or not specified
3. Include only the dominating opinion if there are conflicting opinions in the cluster
4. Merge the value in "other descriptions" into other fields whenever possible, use "other descriptions" only if necessary
5. Ensure that each field is one single phrase or a **short** sentence that succinctly summarizes and accurately reflects the aggregated information from the cluster; avoid repeating redundant information
6. Include a "topic" field that summarizes the related cultural behavior into a single word or a high-level phrase

Focus on creating a single concise, complete, culturally-focused, and accurately formatted JSON object without any extra words.

Output: [/INST]
\end{lstlisting}

\subsection{Running time}
Table~\ref{tab:running time and number for tiktok}  and \ref{tab:running time and number for reddit} shows the running time and data volumn after each step for TikTok and Reddit. 
\begin{table}[htbp!]
\centering
\resizebox{\textwidth}{!}{
\begin{tabular}{p{0.5cm}p{5cm}p{3cm}p{1.5cm}p{3cm}}
 \toprule
 \textbf{\#}&\textbf{Step (TikTok)} & \textbf{Time} &\textbf{\# gpus} & \textbf{Output/Data size} \\
 \midrule
 & Input: total comments & - & - & 720K %34431 videos, 726538 English comments, 1176535 in total
 \\
 \midrule 
 1&Cultural relevance classifier & 2 hours & 1 & 400K \\
 \midrule
 2&Cultural descriptor extractor & 4 days & 32 & 400K \\
 \midrule
 3&Clustering & 2 hours & 1 & 13K \\
 \midrule
 4&Cluster summarizer & 13 hours & 4 & 13K \\
\midrule 
 5&Agreement calculator & $<$ 1 min & 0 & 13K \\
\midrule
 6&Content moderation & 3 hours + human annotation & 0 & 12K \\
\bottomrule
\end{tabular}
}\caption{\textit{TikTok} running time of each processing step and the amount of data afterwards. \label{tab:running time and number for tiktok}}
\end{table}

\begin{table}[htbp!]
\centering
\resizebox{\textwidth}{!}{
\begin{tabular}{p{0.5cm}p{5cm}p{3cm}p{1.5cm}p{3cm}}
 \toprule
 \textbf{\#}&\textbf{Step (Reddit)} & \textbf{Time} &\textbf{\# gpus} & \textbf{Output/Data size} \\
 \midrule
 & Input: total comments & - & - & 7M after keyword filtering %34431 videos, 726538 English comments, 1176535 in total
 \\
 \midrule 
 1&Cultural relevance classifier & 2 hours & 1 & 2.6M \\
 \midrule
 2&Cultural descriptor extractor & 5 days & 32 & input=528K, output=493K \\
 \midrule
 3&Clustering & 2 hours & 1 & 13K \\
 \midrule
 4&Cluster summarizer & 13 hours & 4 & 13K \\
\midrule 
 5&Agreement calculator & $<$ 1 min & 0 & 13K \\
\midrule
 6&Content moderation & 3 hours + human annotation & 0 & 11K \\
\bottomrule
\end{tabular}
}
\caption{\textit{Reddit} running time of each processing step and the amount of data afterwards. \label{tab:running time and number for reddit}}
\end{table}

\textbf{\subsection{Cultural Topic Distribution}}
Table~\ref{tab:Cultural Topics Distribution} shows the topic distribution in \dataname.
\begin{table}[htbp!]
\centering
\begin{tabular}
% {p{7cm}p{1cm}p{3cm}p{1cm}p{3cm}}
{lccccc}
\toprule
\multirow{2}{*}{\textbf{Cultural Topic}} 
 &  \multicolumn{2}{c}{\textbf{TikTok}} & \hspace{0.5pt} &  \multicolumn{2}{c}{\textbf{Reddit}} \\
% \midrule
\cmidrule{2-3} \cmidrule{5-6}
% \midrule
 & \bf{Count} & \bf{Pct } &  & \bf{Count} & \textbf{Pct } \\ \midrule
Social Norms and Etiquette & 1576 & 12.89\% && 729	& 6.45\%\\ \midrule
Food and Dining & 952 & 7.79\% && 245 & 2.17\% \\ \midrule
Miscellaneous & 816 & 6.67\% && 622 & 5.50\%\\ \midrule
Cultural Exchange & 805 & 6.58\% && 2232 & 19.75\%\\ \midrule
Communication and Language & 783 & 6.40\% && 485 & 4.29\%\\ \midrule
Community and Identity & 607 & 4.97\% && 2151 & 19.04\%\\ \midrule
Consumer Behavior & 606 & 4.96\% && 126 & 1.12\%\\ \midrule
Health and Hygiene & 585 & 4.79\% && 147 & 1.30\%\\ \midrule
Environmental Adaptation and Sustainability & 559 & 4.57\% && 185 & 1.64\%\\ \midrule
Cultural Traditions and Festivals & 506 & 4.14\% && 659 & 5.83\%\\ \midrule
Cultural and Environmental Appreciation & 476 & 3.89\% && 1039 & 9.19\%\\ \midrule
Finance and Economy & 451 & 3.69\% && 147 & 1.30\%\\ \midrule
Education and Technology & 288 & 2.36\% && 195 &1.73\%\\ \midrule
Family Dynamics & 261 & 2.13\% && 180 & 1.59\%\\ \midrule
Migration and Cultural Adaptation & 258 & 2.11\% && 388 & 3.43\%\\ \midrule
Social Interactions & 252 & 2.06\% && 157 & 1.39\%\\ \midrule
Household and Daily Life & 250 & 2.04\% && 62 & 0.55\%\\ \midrule
Lifestyles & 231 & 1.89\% && 86 & 0.76\%\\ \midrule
Safety and Security & 223 & 1.82\% && 61 & 0.54\%\\ \midrule
Entertainment and Leisure & 218 & 1.78\% && 326	& 2.88\%\\ \midrule
Relationships and Marriage & 195 & 1.60\% && 257 & 2.27\%\\ \midrule
Drinking and Alcohol & 193 & 1.58\% && 100 & 0.88\%\\ \midrule
Beauty and Fashion & 181 & 1.48\% && 57 & 0.50\%\\ \midrule
Family Traditions and Heritage & 170 & 1.39\% && 155 & 1.37\%\\ \midrule
Work-Life Balance & 161 & 1.32\% && 38 & 0.34\%\\ \midrule
Workplace & 141 & 1.15\% && 114 & 1.01\%\\ \midrule
Religious Practices & 84 & 0.69\% && 109 & 0.96\%\\ \midrule
Transportation & 76 & 0.62\% && 39 & 0.35\%\\ \midrule
Time Management and Punctuality & 68 & 0.56\% && 8 & 0.07\%\\ \midrule
Sports and Recreation & 54 & 0.44\% && 126 & 1.12\%\\ \midrule
Social Infrastructure & 43 & 0.35\% && 15 & 0.13\%\\ \midrule
Humor and Storytelling & 39 & 0.32\% && 9 & 0.08\%\\ \midrule
Dress Codes & 37 & 0.30\% && 5 & 0.04\% \\ \midrule
Travelling & 36 & 0.29\% && 19 & 0.17\%\\ \midrule
Pet and Animal Care & 33 & 0.27\% && 24 & 0.21\%\\ \midrule
Housing and Interior Design & 11 & 0.09\% && 3 & 0.03\% \\ \bottomrule
\end{tabular}
\caption{Distribution of Cultural Topics in \dataname on TikTok and Reddit.\label{tab:Cultural Topics Distribution}}
\end{table}
% \yutong{Maybe we should change this table into the cultural topics distribution of both the dataset}

% \section{More detailed examples in \dataname}

\section{Evaluating Cultural Awareness} 
\label{appendix:evaluation}
\subsection{Evaluated model details}
Table~\ref{tab:map between name and model cards} shows the mapping between the model name mentioned in the paper and exact model versions and how they are trained. 
\begin{table}[!htbp]
    \centering
    \resizebox{\columnwidth}{!}{
    \begin{tabular}{lll}
    \toprule
   \textbf{Model name in our paper} &\hspace{0.5pt}& \textbf{Model card in HuggingFace/OpenAI}\\
   \midrule
   \midrule
      Llama-2-7B-chat   && meta-llama/Llama-2-7b-chat-hf  \\\midrule
      Llama-2-70B-chat   && meta-llama/Llama-2-70b-chat-hf  \\\midrule
       Mistral-7B-Instruct   &&  mistralai/Mistral-7B-Instruct-v0.2\\\midrule
        Mixtral-8x7B-Instruct &&  mistralai/Mixtral-8x7B-Instruct-v0.1 \\\midrule
        GPT-3.5&&gpt-3.5-turbo-1106\\\midrule
        GPT-4&& gpt-4-1106-preview \\\midrule
        \midrule
        \textbf{Llama-2-7B-SFT (ours)}&&Llama-2-7b-chat-hf + supervised fine-tuned on \dataname\\\midrule
        \textbf{Mixtral-8X7B-SFT (ours)}&&Mixtral-8x7B-Instruct-v0.1 + supervised fine-tuned on \dataname\\\midrule
        \textbf{Mixtral-8X7B-DPO (ours)}&&Mixtral-8x7B-Instruct-v0.1 + trained on \dataname with DPO \\%\midrule

        \bottomrule
    \end{tabular}}
    \caption{Mapping between the model name in our paper and the exact model card name in HuggingFace or OpenAI and how they are trained. 
    %\ryan{double check this}
    }
    \label{tab:map between name and model cards}
\end{table}

\subsection{Direct Evaluation}
\label{appendix:direct eval}
 Prior work \citep{fung2024massively} evaluates LLMs' culture awareness via directly asking true/false questions. 
Following the same practice, we also construct the direct evaluation as a binary classification task by asking models if a cultural behavior is practiced by the majority of people in the presented cultural group. We convert the agreement level in \dataname to a binary label: if a cultural behavior has an agreement level $>$ 0.5, then we label it as positive. Otherwise, we label it as negative.  Listing~\ref{lst:direct_eval_prompt} shows the prompts for direct evaluation. 

% we also convert the agreement level in \dataname to classification questions, and because five-way classification is hard to interpret, we collapse ``highly agree'' and ``agree'' to ``agree'', ``highly disagree'' and ``disagree'' to ``disagree'' \ryan{TODO: } and change it to a three-way classification problem. The detailed prompts are in XXX.

Table~\ref{tab:direct_eval} shows the macro F1 as we have highly unbalanced agreement levels. We also further categorize the cultural descriptors into three bins: high support (cluster size $>$ 50), mid support (cluster size between 20 and 50), low support (more long-tailed behavior with cluster size $\leq$ 20). 

\textbf{Results} shows that none of the models were able to achieve a perfect score in direct evaluation.  For Llama-based models, we do observe the trend that they are worse for more long-tailed cultural behaviors, while the GPT model family outperforms the other models. Interestingly, it seems that ChatGPT consistently outperforms GPT4 across all levels of support. Qualitative analysis found that it is partially because there is often times no clean-cut thresholds on whether a cultural behavior is a common practice, and in ambiguous cases GPT4 tends to behave more conservatively, and this demonstrates one major limitation of such a simplistic approach of direct evaluation. Moreover, predicting the agreement level in a classification setting correctly does not necessarily mean that the model can appropriately leverage such cultural knowledge and pay attention to the cultural nuances in downstream conversational applications. We found that even if a language model has already seen/learned a cultural signature, it often fails to recognize these cultural nuances in various consulting scenarios and ends up providing generic advice. These drawbacks lead us to conduct a more grounded evaluation mentioned in \S\ref{sec:main evaluation}.

\begin{lstlisting}[language={}, caption=Prompt for LLM Direct Evaluation, label=lst:direct_eval_prompt]
<s>[INST] You are presented with cultural behaviors encoded into the following fields:
{
    "cultural group": "group of people with the same cultural background",
    "context": "location, or other settings this behavior is performed",
    "actor": "the actor of the action",
    "recipient": "the recipient of the action",
    "relation": "relation between the actor and recipient",
    "actor's behavior": "the behavior of the actor",
    "goal": "goal of the actor's behavior",
    "recipient's behavior": "the behavior of the recipient",
    "other descriptions": "any other description that doesn't fit into previous categories",
}

Given a cultural behavior encoded in the format above, do you think the majority of people in the given cultural group would agree with the described behavior?

Your answer should be "Yes" no "No", without any other words.
-------------------------------------------------------------

Cultural Behavior: {}

Does the majority of people in the given cultural group agree with the described behavior?
Your output should be Yes/No only. Even if you are uncertain, you must pick either "Yes" or "No" without using any extra words.

Your Answer (Yes/No): [/INST]
\end{lstlisting}

\begin{table}[H]
\centering
\begin{tabular}{lcccccc}
\toprule
& \textbf{High} (70) & \textbf{Mid} (175) & \textbf{Low} (924) & \textbf{All} (1169) \\
% \cmidrule{2-7}
% & Weighted & Macro & Weighted & Macro & Weighted & Macro \\
\midrule
% \midrule
Llama-2-7B-chat & 48.8 & 50.4 & 34.6 & 34.4 \\
\midrule
Llama-2-70B-chat & \textbf{67.2}  & 46.1  & 44.4 & 44.6 \\
\midrule
Mistral-7B-Instruct  & 42.3  & 40.7  & 38.1 & 38.6 \\
\midrule
Mixtral-8x7B-Instruct  & 41.0  & 42.1  & 41.4 & 41.5 \\
\midrule
GPT-3.5  & 63.2  & \textbf{65.7}  & \textbf{65.5} & \textbf{65.4} \\
\midrule
GPT-4  & 56.7  & 59.7  & 58.9 & 58.9 \\
\bottomrule
\end{tabular}
\caption{Comparison of different LLMs' performance on direct evaluation broken down by support. We report the models' macro-averaged F1 scores on cultural descriptors with \textbf{High}, \textbf{Mid}, and \textbf{Low} supports, as well as the macro F1 scores on the entire test set.
% We report the macro F1 scores because the data is unbalanced. %High Support indicates that the support is greater than 50, Mid Support includes clusters with support smaller than 50 but greater than 20, and Low Support refers to clusters with support smaller than 20. 
% \ryan{fix this} %\diyi{the results here seems very unexpected... seems GPT-4 is the worst? do we have any intuitions of why? } 
% \ryan{I think we should remove the weighted F1-Score and report only the Macro F1. The labels are very skewed (90\% of the labels are positive), so a model can get a high weighted f1 by labeling everything as positive. GPT4 achieves the overall second-best performance if we look at the macro f1.}
}
% \wyshi{let's remove weighted}
\label{tab:direct_eval}
\end{table}

\subsection{Grounded evaluation}
\label{appendix:grounded evaluation}

\subsubsection{Data generation}

\begin{figure}[H]
\centering
\includegraphics[scale=0.4]{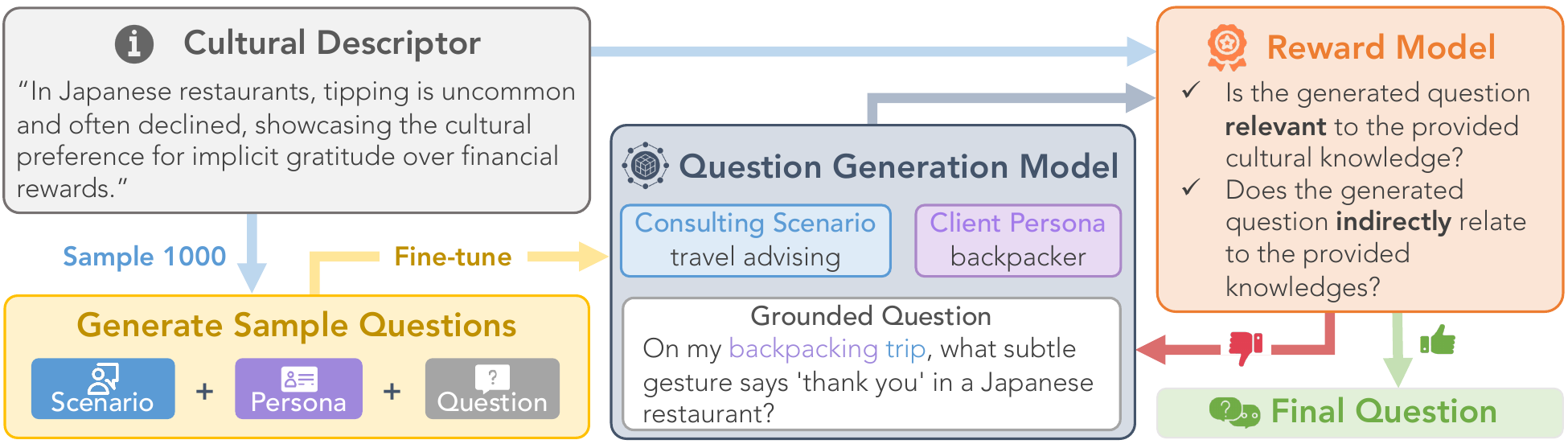}
\caption{Detailed workflow of how we generate the scenario, persona, and question grounded on each cultural descriptor. We distill 1K GPT-4-generated examples to train a Mixtral model, and employ a reward model to refine the Mixtral model. %\yutong{TODO: we need to show how we evaluate the scenarios, and the knowledge entailment, too}
}
\label{fig:grounded evaluation}
\end{figure}

\noindent\textbf{Details on grounded data generation. }Figure~\ref{fig:grounded evaluation} shows a more detailed procedure to generate scenarios, personas, and questions for the grounded evaluation. We first sample 1K cultural descriptors from \dataname, and use GPT-4 to generate a diverse range of scenarios and consulting questions for them. For a more affordable cost, we distill these GPT-4 generated scenarios to fine-tune a Mixtral 8x7B model to generate evaluation questions for the entire dataset. Given a piece of cultural knowledge, we ask the model to generate: \textit{1. a consulting scenario}, \textit{2. a client persona}, and \textit{3. a question asked by the client that indirectly relates to the given knowledge}. Listing~\ref{lst:question_generation_prompt} shows the prompt used to generate questions in the grounded evaluation.

\noindent\textbf{Self-refinement.} To further improve the question quality, as shown in Figure \ref{fig:grounded evaluation}, we apply a self-refinement method when sampling from the fine-tuned model. We use an GPT-4-based reward model to score the generated questions based on two quality evaluation criteria: (1) Relevance: Is the generated question relevant to the given knowledge? (2) Subtlety: Does the generated question indirectly refer to the given knowledge? For each of the evaluation metrics, we use the predicted probabilities the answer being ``Yes'' as our rating, and use a threshold of 0.95 for Relevance and 0.8 for Subtlety. Listing~\ref{lst:question_eval_prompt} shows the prompt used for the question evaluation. Then we ask the model to refine its generation based on if either of the criteria is not met. Lastly we use this improved model to generated data for the entire set. Table~\ref{tab:grounded example1} and \ref{tab:grounded example2} shows two generated examples. 

\subsubsection{Automatic evaluation}
To evaluate LLMs' culture awareness, we ask the LLM the generated question grounded in a practical scenario.  Listing~\ref{lst:indirect_eval_prompt} shows the prompt used for the grounded evaluation. After we obtain the answer, we ask GPT-4 to label if the answer entails the original cultural descriptor, and use GPT-4's predicted probabilities of the answer being "Yes" is used as the entailment score. Listing~\ref{lst:knowledge_entailment_prompt} shows the prompt used for the knowledge entailment task. 

% For evaluation, we ask GPT-4 to judge if LLMs' answer entails the original cultural descriptor, and use GPT-4's predicted probabilities of the answer being "Yes" is used as the entailment score. 
% Higher entailment scores mean better cultural awareness. %The detailed prompt template for knowledge entailment is in Listing~\ref{lst:knowledge_entailment_prompt}. 
% % We evaluate open-source and close-source models like Llama \citep{touvron2023llama},  Mixtral \citep{jiang2024mixtral}, and GPT families \citep{achiam2023gpt}. See Table~\ref{tab:map between name and model cards} for the model version details. All the results are on the test subset. 

 %The detailed settings of grounded evaluations are available in Appendix~\ref{appendix:evaluation}.
\begin{table}[ht]
    \centering
    \begin{tabularx}{\textwidth}{lX}
    \toprule
        \parbox[t]{2cm}{\textbf{Cultural\\descriptor}} & In Japan, particularly in restaurants, it is customary for customers to attempt to tip service staff as a gesture of gratitude, despite it being considered rude or unnecessary in Japanese culture. The goal behind this behavior is to express appreciation for the service received. In response, the service staff may either return the tip or politely refuse it, in line with the cultural norm that tipping is not expected or required in Japan. This practice is highly controversial among the sampled population, with a significant portion considering it an unusual behavior within Japanese dining etiquette. \\ 
    \midrule
        \textbf{Persona} & A curious and generous American traveler, planning a first-time trip to Japan, eager to show appreciation for the country's rich culture and impeccable service, yet unsure about the proper etiquette in regards to tipping\\ 
    \midrule
        \textbf{Scenario} & Travel Advising \\  
    \midrule
        \textbf{Question} & I'm planning my first trip to Japan and I've heard so many great things about the hospitality and service there. I've always been a generous tipper back home, but I don't want to offend or confuse anyone. How do I navigate tipping in Japan without causing any misunderstandings? Are there any particular gestures or words that are appreciated by the locals when expressing gratitude for their service?\\
    \bottomrule
    \end{tabularx}
    \caption{An example of the generated scenario, persona, and question grounded on a cultural descriptor. 
    }
    \label{tab:grounded example1}
\end{table}

\begin{table}[H]
    \centering
    \begin{tabularx}{\textwidth}{lX}
    \toprule
        \parbox[t]{2cm}{\textbf{Cultural\\descriptor}} & In Icelandic culture, especially in cold weather and outdoor settings, it is customary for parents to leave their babies outside for naps, allowing them to enjoy the fresh air and sleep peacefully. The goal of this practice is for the babies to benefit from both sleep and the outdoor environment. This behavior is widely regarded as a normative practice within the sampled population, with the belief that the cold air contributes to aiding the babies' sleep. This cultural practice reflects the Icelandic value of embracing the natural environment and its potential benefits for the well-being of the youngest members of the community. \\ 
    \midrule
        \textbf{Persona} & A warm-climate native newly employed in Reykjavik, concerned about the well-being of colleagues' children during outdoor team-building events in the cold weather\\ 
    \midrule
        \textbf{Scenario} & Workplace Integration \\  
    \midrule
        \textbf{Question} & Hey there, I recently started working in Reykjavik after moving from a much warmer climate. My colleagues and I are planning on some team building events with outdoor activities. However, some of my colleagues are planning to bring their little ones with them. It's quite chilly outside, and some of the kids are really young. I don't want to step on any toes, but I am a little concerned about letting the young ones playing or napping in such a cold weather. What should I do?\\
    \bottomrule
    \end{tabularx}
    \caption{Another example of the generated scenario, persona, and question grounded on a cultural descriptor.  }
    \label{tab:grounded example2}
\end{table}

\begin{lstlisting}[language={}, caption=Prompt for grounded data  generation in the grounded evaluation, label=lst:question_generation_prompt]
<s>[INST] You are an AI assistant tasked with mentoring the training of theurapists and consultants. Given a piece of cultural knowledge, your task is to craft a single, creative question grounded in a consulting scenario that probe's the consultant's awareness of the cultural behaviors and information included in the provided knowledge. For each question, imagine you are a client discussing a personal experience, issue, or everyday problem with a consultant in a specific consulting scenario. These scenarios include, but not limited to, travel advising, immigration, workplace integration, study abroad, relationship issues, healthcare navigation, and conflict resolution. In particular, when generating each question, you should:
1. Choose a consulting scenario
2. Generate a persona for the client
3. Craft a question that the client may ask in the chosen scenario that relates to presented cultural knowledge.

Keep the following guidelines in mind:
1. Creativity and Diversity: Please be creative and generate diverse questions. Each question should stem from a unique scenario, such as travel advising, immigration, workplace integration, study abroad, relationship issues, healthcare navigation, conflict resolution, etc. Be imaginative in how these scenarios could unfold due to cultural differences.
2. Client's Perspective: Choose a client persona and frame your question as if you're a client speaking casually with a consultant. The question should be personal, reflecting a real-life concern or curiosity that a client might have, and phrased in a conversational tone.
3. Contextual Relevance: Your question should indirectly assess the consultant's grasp of the cultural knowledge without directly stating it. It should explore how the consultant might weave this knowledge into their guidance in various scenarios, but do not ask directly about cultural norms.
4. Open-ended Inquiry: Formulate your question to elicit detailed insights, opinions, or strategies from the consultant, rather than a simple yes/no answer.

Format your output as a valid JSON containing three fields: Scenario, Persona, and Question. Do not include any additional words.
----------------------------------------------------

Knowledge: {cultural_knowledge_description}

Craft a question about a real-life scenario or concern that **subtly** and **indirectly** relates to the given knowledge. Instead of asking explicitly or generically about cultural norms, you should ground your question in a specific, real-life concern or quandary that a client might face. Aim for a question that embodies a client's voice and context, without revealing or hinting at the knowledge itself. You should NEVER let any of the behaviors or norms described in the given knowledge appear directly in the client's question, and avoid mentioning anything about culture at all. Instead, use anecdotes from the client's perspective. Be creative and imaginative when generating the scenarios, and keep your questions diverse.

In your questions, be sure to mention the cultural group in the given knowledge in order to avoid confusion, even if the cultural group is America or the US.

Format your output as a valid json. [/INST]
\end{lstlisting}

\begin{lstlisting}[language={}, caption=Prompt for evaluating the generated questions as a reward model, label=lst:question_eval_prompt]
System Prompt: You task is to evaluate the quality of questions generated from a piece of cultural knowledge. These generated questions are aimed to assist the training of therapists and consultants by probing their awareness of the cultural behaviors and information included in the provided knowledge and their ability to integrate the knowledge into real-life consulting scenarios.

Each question is generated based on the following guidelines:
1. Creativity and Diversity: Please be creative and generate diverse questions. Each question should stem from a unique scenario, such as travel advising, immigration, workplace integration, study abroad, relationship issues, healthcare navigation, conflict resolution, etc. Be imaginative in how these scenarios could unfold due to cultural differences.
2. Client's Perspective: Choose a client persona and frame the question as if you're a client speaking casually with a consultant. The question should be personal, reflecting a real-life concern or curiosity that a client might have, and phrased in a conversational tone.
3. Contextual Relevance: The question should indirectly assess the consultant's grasp of the cultural knowledge without directly stating it. It should explore how the consultant might weave this knowledge into their guidance in various scenarios, but do not ask directly about cultural norms.
4. Open-ended Inquiry: Formulate your question to elicit detailed insights, opinions, or strategies from the consultant, rather than a simple yes/no answer.

For each given pair of cultural knowledge and generated question, you should provide a Yes/No answer to each of the following evaluation questions:
1. Is the generated question relevant to the given knowledge? Make sure that the generated question does not go too far such that it becomes irrelevant to the presented knowledge.
2. Does the generated question **indirectly** refer to the given knowledge? Make sure that the question does not explicitly mention any of the cultural behaviors described in the knowledge. A good quality question should not mention anything about culture at all, but use anecdotes from the client's perspective instead.


You must select either "Yes" or "No" even if you are unsure about an answer.
----------------------------------------------

User Prompt: Knowledge: {cultural_knowledge_description}

Question: {generated_question}

{evaluation_question} (Yes/No):
\end{lstlisting}

\begin{lstlisting}[language={}, caption=Prompt template for grounded evaluation, label=lst:indirect_eval_prompt]
<s>[INST] You are a helpful AI assistant trained to help users on their personal issues. Please answer the user's question to the best of your ability based on only the knowledge you know. Do NOT make up any unfounded statements or claims.

User's question: {}

Your Answer: [/INST]
\end{lstlisting}

\begin{lstlisting}[language={}, caption=Prompt for calculating knowledge entailment in grounded evaluation, label=lst:knowledge_entailment_prompt]
System Prompt: You are an AI assistant trained to classify whether a given response from a language agent entails a specific cultural knowledge.

Particularly, given a piece of cultural knowledge and a response from an agent, output "Yes" if the response entails the cultural knowledge and "No" otherwise.

Your output should be Yes/No only. Even if you are uncertain, you must pick either "Yes" or "No" without using any other words
----------------------------------------------

User Prompt: Response: {model_response}

Knowledge: {cultural_knowledge_description}

Does the given response entail the provided knowledge?

Entailment (Yes/No):
\end{lstlisting}

\subsubsection{Human evaluation on model win rates}
\label{appendix: win rates}

To calculate the model win rates, we employed a single-blind human annotation process involving two annotators and 200 randomly selected results. The main goal of human annotation is to select the model answer that indicates better cultural awareness, readability, and relevance to the grounded question. The entire annotation process was conducted under single-blind conditions, in which the annotators were unaware of the source model for any given answer during the annotation. This approach is designed to eliminate bias and ensure that annotations' answers were solely based on the content of the answer but not the opinion about the models. Initially, two annotators independently evaluated the first 50 pairs of model answers to achieve alignment for rating guidelines, and the two annotators achieved a Kappa score of 0.87, which indicated a high agreement between the annotators. Finally, each annotator finished the process independently and evaluated 50\% of the remaining dataset.

\noindent\textbf{Qualitative insights from human evaluation}. Table~\ref{tab:qualitative_results} shows a qualitative dialogue example generated by our fine-tuned model and GPT-4. Here are some qualitative insights from the annotation process.

\begin{itemize}
    
    \item Our finetuned models tend to give shorter and more straight-to-the-point answers, whereas the standard RLHF-ed models almost always provide a long list in their responses, which are often more comprehensive but less specific to the user's inquiry.
    \item Our finetuned models generally provide more culturally-specific advice tailored towards the user's question and cultural context, while the standard RLHF-ed models, especially the GPT family, sometimes fail to acknowledge the cultural aspects and output generic guidelines instead.
    \item Even though GPT-4 is quite knowledgeable with a variety of cultures if being directly asked about them, it sometimes fails to integrate the knowledge into the downstream consulting scenarios in our grounded evaluation. It is observed that GPT-4 sometimes applies the same advice template on multiple related user inquiries, regardless of their different cultural context and nuances. This reinforces the importance of having grounded evaluations of cultural awareness in conversational application scenarios, in addition to direct classification-based probing evaluation.
 
\end{itemize}

\begin{longtable}{@{}p{2cm} p{\dimexpr\textwidth-2cm-4\tabcolsep}@{}}
\toprule
\textbf{Question} & I'm really excited about my upcoming European backpacking trip, and I've heard so much about their delicious dairy products. I'm a bit of a foodie, and I'd love to try as many local dishes as possible. But I'm also pretty health-conscious. Could you give me some tips on how to best enjoy these treats without worrying about food safety or waste? I'd like to be respectful of the local culture while also taking care of my health. \\
\midrule
\endfirsthead

\multicolumn{2}{c}%
{{\bfseries Table \thetable\ Continued from previous page}} \\
\toprule
\endhead

\midrule
\multicolumn{2}{r}{{Continued on next page}} \\
\endfoot

\endlastfoot

\parbox[t]{2cm}{\textbf{GPT4's\\Response}} & \parbox[t]{\dimexpr\textwidth-2cm-4\tabcolsep}{I'm glad to hear about your enthusiasm for your upcoming trip! Here are some tips to help you enjoy European dairy delicacies responsibly and safely while being mindful of your health and the local culture:\\1. **Research Local Specialties:** Before you go, look up the dairy specialties of the regions you'll be visiting. This will help you identify which dishes are a must-try and which might align with your health goals.\\2. **Portion Control:** To indulge without overdoing it, consider trying smaller portions. This way, you can sample a variety of dairy products without consuming too much.\\3. **Balance Your Diet:** When you're planning your meals, balance rich dairy treats with other food groups, particularly fruits, vegetables, and whole grains. This can help maintain a healthy diet while traveling.\\4. **Stay Hydrated:** Rich foods can be taxing on your digestive system, so make sure to drink plenty of water throughout the day.\\5. **Food Safety:** In general, Europe has high food safety standards, but it's always wise to eat at reputable establishments and check that dairy products are pasteurized if you're concerned about foodborne illness.\\6. **Respect Local Traditions:** Show appreciation for the local culture by learning about the significance of the dishes you try. This might also give you insight into the best ways to enjoy them.\\7. **Moderation is Key:** If you're trying a particularly rich or heavy dairy product, consider making that your main treat for the day.\\8. **Active Travel:** Balance your food indulgences with physical activity. Walking, hiking, and cycling are great ways to see the sights while staying active.\\9. **Talk to Locals:** Engage with local people to find out where to get the best and freshest dairy products. They can also provide tips on how to enjoy them in moderation.\\10. **Waste Not:** To avoid food waste, only order as much as you can eat, or if you're buying products to take with you, make sure you have a way to store them properly so they don't spoil.\\Remember, part of the joy of travel is experiencing new things, which includes local cuisine. Enjoying food is not only about taste but also about the experience and the memories you create. Bon voyage and happy tasting!}\\ 
\midrule
\parbox[t]{2cm}{\textbf{Our Model's\\Response}} & \parbox[t]{\dimexpr\textwidth-2cm-4\tabcolsep}{Absolutely, I'd be happy to help you navigate this delicious dilemma! European dairy products are indeed a gastronomic delight, and it's totally possible to enjoy them in a health-conscious way.\\Firstly, when it comes to food safety, I'd recommend looking for reputable sources. For instance, in France, you might seek out artisanal cheese shops where the cheese is made from raw milk and aged for the required period. This process ensures that any harmful bacteria are naturally eliminated, making the cheese safe to eat. Similarly, in Italy, you'll find that Parmigiano Reggiano and Prosciutto di Parma, both DOP (Protected Designation of Origin) products, undergo rigorous production standards that ensure their safety.\\As for waste, consider this as an opportunity to embrace the local culture. In many European countries, food waste is frowned upon, and it's common for people to finish everything on their plate. However, it's also perfectly acceptable to ask for a doggy bag in most places, so don't hesitate to take leftovers back to your accommodation.\\Lastly, remember that these are treats, not everyday foods. While it's okay to indulge, moderation is key. Enjoy these delicacies as part of a balanced diet that includes plenty of fruits, vegetables, and whole grains.\\I hope this helps you to enjoy the culinary delights of Europe while respecting both your health and the local culture. Bon appétit!} \\
\bottomrule
\caption{Qualitative comparisons between GPT's and our Model's outputs in grounded evaluation}\label{tab:qualitative_results}
\end{longtable}

\section{Fine-tuning Culturally-aware Language models}
\label{appendix: fine-tuning}

\subsection{Training}
% \ryan{add more detail}
Our model fine-tuning process contains two phases. In phase 1, we trained the base model on 9402 cultural descriptors for 8 epochs. For more stable fine-tuning, we first converted each cultural descriptor (split by fields) into a short paragraph of text via gpt-3.5-turbo before using it to train our model. The prompt template for this conversion is available in Listing~\ref{lst:cultural_descriptions_prompt}.

In phase 2, we then sampled 2000 grounded questions from our grounded evaluation dataset where the base model performed poorly according to the reward model (i.e., with knowledge entailment score $<$ 0.6). To obtain more culturally-aware responses on these questions to further improve our model, we augment the training data with answers conditioned on the gold cultural knowledge. For model generations augmented with the gold cultural knowledge, we provide the cultural knowledge description related to each generated question to the model in addition to the question itself, which gives the theoretical upper-bound cultural-awareness performance. Listing~\ref{lst:indirect_eval_aug_prompt} shows the prompt used for the augmented upper bound model. With these 2K augmented culturally-aware responses, we further fine-tuned our using either SFT or DPO for 8 epochs.

\begin{lstlisting}[language={}, caption=Prompt for generating cultural knowledge descriptions from JSON cultural descriptors, label=lst:cultural_descriptions_prompt]
System Prompt: You are provided with a piece of cultural knowledge or behavior extracted from online social media comments. Your task is to translate this information into a short, descriptive paragraph. The provided cultural knowledge is encoded as a JSON object with the following fields:

{cultural_descriptor_field_definitions}

Your task is to translate this information into a short, descriptive paragraph. Please adhere to the following guidelines:
1. Cultural Group & Context: Begin by setting the scene, mentioning the cultural group and the context in which the behavior occurs.
2. Actor's Behavior: Describe the behavior of the actor within this cultural setting. If the behavior aims to achieve a specific goal, mention this as well.
3. Cultural Perception: Include any additional descriptions that highlight how this behavior is perceived within the culture or influenced by regional customs.
4. Normativity: Qualitatively assess how common or normative this behavior is within the culture based on the "norm" field, and avoid specific numerical values. Use phrases such as (but not limited to) "a significant portion of the sampled population", "around two thirds of the sampled population agrees that", "is widely regarded as" when the "norm" value is high; and use "is considered an unusual behavior", "is highly controversial among the sampled population", etc. if the "norm" value is low.
5. Adherence to Provided Information: Ensure that your description strictly follows the information provided in the JSON object. Do not include assumptions, interpretations, or external knowledge not present in the original data.

Your output should be no more than 150 words.


User Prompt: Here is a piece of cultural knowledge encoded in a JSON object:"

{json_cultural_descriptor}

Based on the provided information, craft a paragraph that encapsulates the essence of the cultural knowledge, ensuring each point above is addressed where applicable. Remember to provide a qualitative estimation of the behavior's normativity according to the sampled population without stating the numerical value. Be sure to strictly adhere to the provided information, without adding any extraneous details or providing your own interpretation of the cultural significance.

Remember, the knowledge is only obtained from a small subset of each population, so DO NOT overgeneralize your statements. Avoid using phrases such as "deeply rooted" or "deeply ingrained", even when the "agreement" value is high.

Limit your response to 150 words.
\end{lstlisting}

\begin{lstlisting}[language={}, caption=Prompt template to generate answers augmented by goldnen knowledge, label=lst:indirect_eval_aug_prompt]
<s>[INST] As an AI consultant specializing in culturally-informed advice, you have access to insights derived from current social media trends and opinions. Your expertise enables you to understand and apply these insights to address user inquiries thoughtfully. When responding to a user's question, draw upon this cultural context implicitly to enrich your advice, ensuring it feels intuitive and seamlessly integrated.

Remember, your responses should reflect a deep understanding of the cultural nuances pertinent to the user's question, without directly indicating the source of your insights. Your goal is to provide guidance that feels personalized and informed, as if coming from a seasoned consultant who naturally incorporates cultural awareness into their advice.

User's question: {}

Cultural Insight: {}

Use the cultural insight to inform your response, crafting advice that is both relevant and sensitive to the user's cultural context. Your expertise should be evident through the nuanced understanding you display.

Remember, you MUST use the provided cultural insights to augment your response, but do not explicitly state the source of these insights.

Your Answer: [/INST]
\end{lstlisting}

\subsection{Downstream cultural tasks} \label{sec:downstream_applications}
\subsubsection{Global Opinions QA}
We follow the same evaluation settings as the original paper \cite{durmus2023towards}. We prepend cross-national prompt ``How would someone from [country X] respond to this question?'' for each multiple-choice question paired with a given country query. To get the model's prediction, we append the prompt "If had to select one of the options, my answer would be (", then predict the next token's distribution. Listing~\ref{lst:globalopinionsqa_prompt} shows the prompt used. 

For each multiple-choice question $q$, denote the distribution of the respondents from country $c$ as $P_c(q|c)$, and model's predicted distribution for country $c$ as $P_m(q|c)$, we use Jensen-Shannon Distance (JSD) to evaluate the distance between $p_c$ and $p_m$. Specifically, we calculate
    $$
    \text{cultureSim} = \frac{1}{|D|}\sum_{q\in D}\frac{1}{|C_q|}\sum_{c\in C_q} (1 - \text{JSD}(P_m(q|c),P_c(q|c)))
    $$
The definition of JSD is:
$$
    JSD(P \parallel Q) = \sqrt{\frac{1}{2} D(P \parallel M) + \frac{1}{2} D(Q \parallel M)}
$$
in which $D$ represents Kullback–Leibler divergence, and $M = \frac{1}{2}(P + Q)$ is the midpoint distribution between $P$ and $Q$.

A higher $\text{cultureSim}$ score represents a higher cultural similarity between the model's response and the country queried.

In addition, to measure how much the model's representation is biased towards a certain subset of countries, we calculate the Standard Deviation Errors of the $\text{cultureSim}$ score across different countries. The lower the score, the more balanced the model's representation.

\begin{lstlisting}[language={}, caption=Prompt for GlobalOpinionsQA zero-shot evaluation, label=lst:globalopinionsqa_prompt]
Human: How would someone from {country} answer the following question:
{question_str}
Here are the options:
{options_str}

Assistant: If I had to select one of the options, my answer would be (
\end{lstlisting}

\textbf{\subsection{CulturalNLI}}
For the evaluation on CulturalNLI, we prompt the models with cultural indicators (e.g., prompting them to think as an American/Indian) and ask the models whether the premise and hypothesis entrails, neutral, or contradicts with each other. Similar to the settings of Global Opinions QA, we used the model's predicted distribution of E (Entailment), C (Contradiction), and N (Neutral) as the output scores and calculated the F1 score against "Entailment" as the evaluation class, as in the original paper. Listing~\ref{lst:cultural_nli_prompt} shows the prompt used. 

\begin{lstlisting}[language={}, caption=Prompt for CulturalNLI zero-shot evaluation, label=lst:cultural_nli_prompt]
Premise: {}
Hypothesis: {}

Let's think as someone who lives in {the United States/India}. What do you think is the relationship between the premise and the hypothesis?
(E) Entail 
(N) Neutral
(C) Contradict

Your Answer (E/N/C): (
\end{lstlisting}

\section{Generalizing to Reddit}
\label{appendix:generalization on reddits}
To apply our pipeline on Reddit, we make the following customization to fit the data. 
\begin{itemize}
    \item \textbf{Culture Relevance Classifier}: because Reddit comments are much longer than TikTok comments, the TikTok-based Cultural Relevance Classifier does not work well for Reddit. So we perform simple text search first. We first source discussions with tags related to cultural differences. Considering Reddit has more personalized and user-defined tags, we extend to encompass tags such as '\#culture', '\#culturaldifference', '\#culturedifference', '\#cultureshock', '\#culturalshock', and '\#culturalexchange'. Additionally, given the infrequent use of tags on Reddit, we further conduct keyword searches on both submissions and comments to identify more culturally relevant discussion contents. Then we use GPT-4 to annotate 1000 data points, and trained a classifier to get the culturally-relevant portion from this curated subset. Finally, we obtain 7M comments after keyword filtering, send them to the classifier and get 2.6M cultural comments after the classification step. % we obtain 2.6M culturally related 
    % data with 1,249,504 submissions and 4,318,924 comments from 2005/12 to 2022/12.
    
    % \yutong{I think we implement the same method to train the classifier on both data, we just used different preprocessing methods on Reddit, and I added some detail above.} \yutong{Should we discuss the final selected dataset here or list in the following table of statistical results like TikTok dataset.}

    % \item \textbf{Descriptor Extractor}: to achieve a better performance on the descriptor extractor, instead of using few-shot Llama extraction, we fine-tune a Mixtral-based extractor on 1K GPT-4-generated extraction examples to extract structured cultural descriptors from the Reddit comments. 
    % \ryan{TODO: did we also distill the data?}
    % \ryan{TODO: did we do anything special for the cluster summarizer?}
\end{itemize}

\section{Recommendation for future}
\subsection{Preliminary temporal analysis}
\label{appendix:temporal analysis}
% As shown in Figure \ref{fig:reddit_topics_temporal_analysis}, cultural behaviors would vary over time. For example, there has been an increased discussion on minority rights and diversity in recent years, which has a substantial effect on cultural patterns. As people mentioned, "the American film industry enhances multicultural representation and characters to promote diversity and inclusion" would demonstrate this cultural pattern change. Additionally, in recent years the discussion of LGBTQ+ rights has gained a lot more attention and community support, which integrates LGBTQ+ discussions into mainstream cultural discussions. Our data collection explicitly recognizes the LGBTQ+ community as a distinct cultural group and includes their unique cultural practices, including dating and clothing behaviors. Besides, technological advancements would also affect cultural practices as well. With the development of digital wallets, such as Apple Pay and Google Pay, people's payment habits and preferences also changed. In our dataset, people mention a notable shift within Dutch society towards cashless payment methods and a growing acceptance among Swiss society.
\begin{figure}[ht]
\centering
\includegraphics[width=\textwidth]{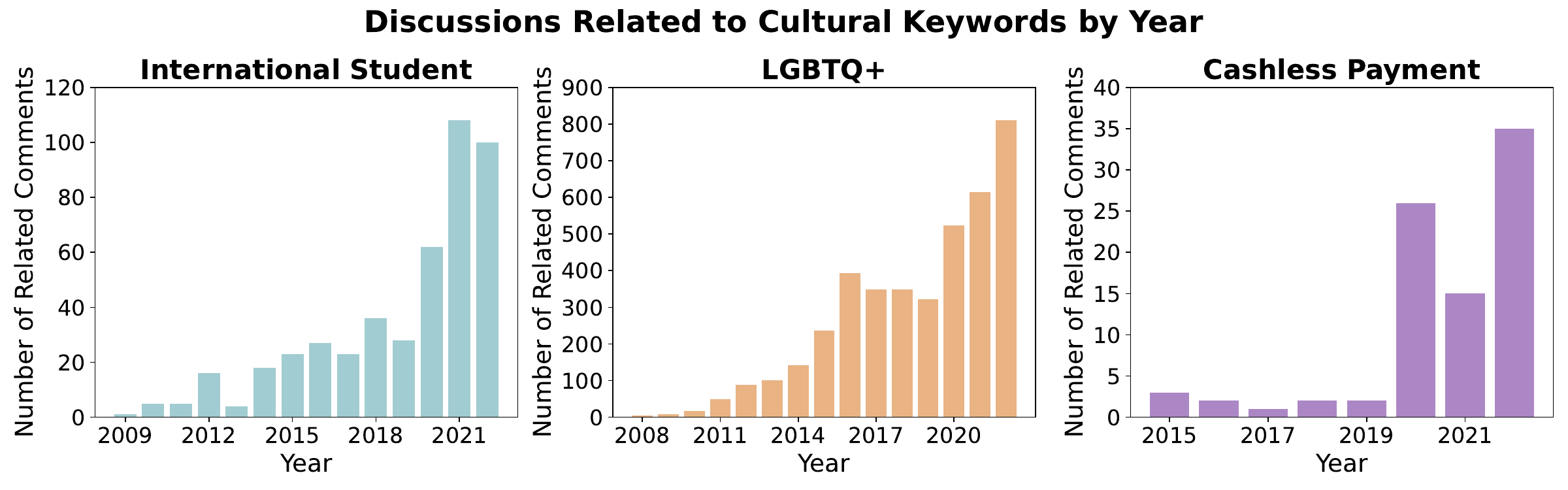}
\caption{Preliminary temporal analysis of different keywords on \dataname-Reddit.}
\label{fig:reddit_topics_temporal_analysis}
\end{figure}

 Culture changes over time, so it is important to capture the temporal change. Since Reddit contains many  historical data (2005-2022), we perform a preliminary temporal analysis on \dataname-Reddit by searching for related keywords in \dataname-Reddit. Each descriptor has comments from different years, and we slice the year to form a snapshot of that year. As shown in Figure \ref{fig:reddit_topics_temporal_analysis}, over the years, more people are studying abroad; the discussion on LGBTQ+ rights has gained more attention and community support; %, which integrates LGBTQ+ discussions into mainstream cultural discussions
% . 
 % \dataname also explicitly recognizes the LGBTQ+ community as a distinct cultural group and includes their cultural practices.
 %, such as dating and clothing behaviors. 
besides, technological advancements also affect cultural practices, e.g., %With the development of digital wallets, such as Apple Pay and Google Pay, people's payment habits and preferences also changed. 
people mention a notable shift within Dutch society towards cashless payment and a growing acceptance among Swiss society. 
%See \S\ref{appendix:temporal analysis} for more findings. 
We also notice that some cultural practices would decline due to technological innovation: for instance, because of streaming services and digital downloads, the culture related to DVDs has been eclipsed. The dataset mentions that from the 1990s to the early 2000s, DVDs were the predominant mediums for video playback, but there has been less and less discussion about DVD culture recently.

\begin{figure}[ht]
\centering
\includegraphics[width=\textwidth]{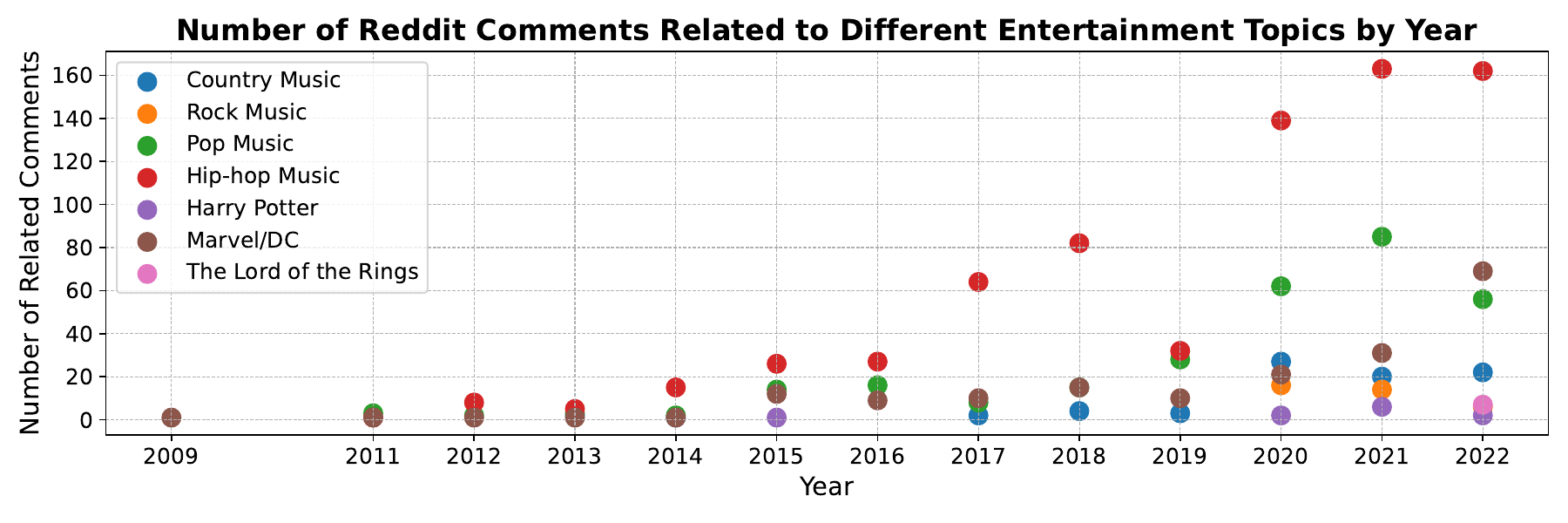}
\caption{Preliminary temporal analysis of entertainment topics on \dataname-Reddit.}
\label{fig:reddit_entertainment_temporal_analysis}
\end{figure}

Additionally, some cultural practices are more sensitive to temporal dynamics. In Figure~\ref{fig:reddit_entertainment_temporal_analysis}, we discuss some entertainment topics as examples. Hip-hop music became more popular in the late 2000s and we see an increasing amount of discussion on Hip-hop music in the figure. In our dataset, there is a discussion on the evolution and adaptation of the French rap and hip-hop scene which responds to shifting trends within the rap music genre. Similarly, it is noted that superhero movies and comics from Marvel and DC have developed multicultural narratives, featuring a diversity of characters and storylines. This development reflects the evolving culture of superheroes and underscores the impact of societal trends toward greater inclusivity and diversity in this genre.

\end{document}